%% file: main.tex
\documentclass[paper,onecolumn,twoside]{geophysics}

\date{} 

\usepackage{enumitem}

\usepackage{url}

\usepackage{hyperref} 

\renewcommand{\figdir}{Fig}

\usepackage{amssymb}
\usepackage{mathrsfs}
\usepackage{amsmath}
\usepackage{algorithm,algorithmic}
\usepackage{url}
\usepackage{multirow}
\usepackage{threeparttable}
\makeatletter
\newcommand{\ALOOP}[1]{\ALC@it\algorithmicloop\ #1%
  \begin{ALC@loop}}
\newcommand{\ENDALOOP}{\end{ALC@loop}\ALC@it\algorithmicendloop}

\usepackage{setspace}
\usepackage[skip=2pt, font=normalsize]{caption}
\makeatother

\begin{document}

\onecolumn 

\begin{description}[labelindent=-2cm,leftmargin=2cm,style=multiline]

\item[\textbf{Citation}]{Z. Long, Y. Alaudah, M. A. Qureshi, Y. Hu, Z. Wang, M. Alfarraj, G. AlRegib, A. Amin, M. Deriche, S. Al-Dharrab, and H. Di, “A comparative study of texture attributes for characterizing subsurface structures in seismic volumes,” Interpretation, vol. 6, no. 4, pp. T1055-T1066, Nov. 2018.}

\item[\textbf{DOI}]{\url{https://doi.org/10.1190/INT-2017-0181.1}}

\item[\textbf{Review}]{Date of publication: 24 October 2018}

\item[\textbf{Data and Codes}]{\url{https://ghassanalregib.com/}}

\item[\textbf{Bib}] {@article\{long2018comparative,\\
  title=\{A comparative study of texture attributes for characterizing subsurface structures in seismic volumes\},\\
  author=\{Long, Z. and Alaudah, Y. and A. Qureshi, M. and Hu, Y. and Wang, Z. and Alfarraj, M. and AlRegib, G. and Amin, A. and Deriche, M. and Al-Dharrab, S. and Di, H.\},\\
  journal=\{Interpretation\},\\
  volume=\{6\},\\
  number=\{4\},\\
  pages=\{T1055--T1066\},\\
  year=\{2018\},\\
  month=\{Nov.\},\\
  publisher=\{Society of Exploration Geophysicists and American Association of Petroleum Geologists\}\}
} 


\item[\textbf{Copyright}]{\textcopyright 2018 SEG and AAPG. Personal use of this material is permitted. Permission from SEG and AAPG must be obtained for all other uses, in any current or future media, including reprinting/republishing this material for advertising or promotional purposes, creating new collective works, for resale or redistribution to servers or lists, or reuse of any copyrighted component of this work in other works.}

\item[\textbf{Contact}]{\href{mailto:zhiling.long@gatech.edu}{zhiling.long@gatech.edu}  OR \href{mailto:alregib@gatech.edu}{alregib@gatech.edu}\\ \url{https://ghassanalregib.com/} \\ }
\end{description}

\thispagestyle{empty}
\newpage
\clearpage
\setcounter{page}{1}

\title{A comparative study of texture attributes for characterizing subsurface structures in seismic volumes}

\renewcommand{\thefootnote}{\fnsymbol{footnote}}

\address{
\footnotemark[1] Center for Energy and Geo Processing (CeGP) at Georgia Tech and KFUPM,
School of Electrical and Computer Engineering,
Georgia Institute of Technology, Atlanta, GA 30332-0250
\footnotemark[2] Center for Energy and Geo Processing (CeGP) at Georgia Tech and KFUPM,
Department of Electrical Engineering,
King Fahd University of Petroleum and Minerals (KFUPM), Dhahran 31261, Saudi Arabia
}
\author{Zhiling Long\footnotemark[1], Yazeed Alaudah\footnotemark[1], Muhammad Ali Qureshi\footnotemark[2], Yuting Hu\footnotemark[1], Zhen Wang\footnotemark[1], Motaz Alfarraj\footnotemark[1], Ghassan AlRegib\footnotemark[1], Asjad Amin\footnotemark[2], Mohamed Deriche\footnotemark[2], Suhail Al-Dharrab\footnotemark[2], and Haibin Di\footnotemark[1]}

\lefthead{Long et al.}
\righthead{Texture for subsurface characterization}

\begin{abstract}
\input{abstract.tex}
\end{abstract}

\section{Introduction}
\label{sec:intro}
\input{introduction.tex}

\section{Texture Attributes}
In this section, we discuss the attributes of interest in this study, divided into two groups. The first group is the traditional attributes including the GLCM-based attributes and the semblance, which are widely used in seismic interpretation. The second group is the local descriptors, which are the newer techniques proposed in the image processing literature in recent years. They are introduced in the order of LBP, LBP variants, and LRI.
\subsection{Traditional Attributes}
\subsubsection{GLCM-based Attributes}
\label{sssec:glcm}
\input{glcm.tex}

\subsubsection{Semblance}
\label{sssec:semblance}
\input{semblance.tex}

\subsection{Local Descriptors}
\subsubsection{Local Binary Pattern}
\label{sssec:lbp}
\input{lbp.tex}

\subsubsection{LBP Variants}
\label{sssec:lbp_variants}
\input{lbp_variants.tex}

\subsubsection{Local Radius Index}
\label{sssec:lri}
\input{lri.tex}

\section{Computational Seismic Volume Labeling}
\label{sec:labeling}
\input{labeling.tex}

\section{Results}
\label{sec:results}
\input{results.tex}

\section{Conclusion}
\input{conclusion.tex}

\section{ACKNOWLEDGMENTS}
\input{acknowledgements.tex}

\bibliography{main}
\bibliographystyle{seg}  

\end{document}

%% file: abstract.tex
In this paper, we explore how to computationally characterize subsurface geological structures presented in seismic volumes using texture attributes. For this purpose, we conduct a comparative study of typical texture attributes presented in the image processing literature. We focus on spatial attributes in this study and examine them in a new application for seismic interpretation, i.e., seismic volume labeling. For this application, a data volume is automatically segmented into various structures, each assigned with its corresponding label. If the labels are assigned with reasonable accuracy, such volume labeling will help initiate an interpretation process in a more effective manner. Our investigation proves the feasibility of accomplishing this task using texture attributes. Through the study, we also identify advantages and disadvantages associated with each attribute.

%% file: introduction.tex
Texture patterns are commonly observed in images of natural scenes \cite[]{GW06}. They have been studied extensively in the image processing literature. In recent years, texture characterization has become an active research area in texture image analysis. In this area, several useful techniques have been developed to extract attributes from a texture pattern that capture the unique spatial distribution of the pixel intensities \cite[]{Haralick1973,ojala2002multiresolution,guo2010completed,Liu201286,Zhai2013:LRI,Zhai2014:LRI-R, hu2016cldp}. These attributes have been proved effective for various texture image analysis tasks such as classification (categorizing textures into groups of different visual appearances), segmentation (identifying boundaries between different textures in a given image), retrieval (finding images with a texture pattern matching a given image), tracking (following changes of spatial location of a certain texture pattern across different images), etc.

Given that seismic data resembles natural texture images in appearance, we believe that we can analyze them from an image processing perspective. From this perspective, a seismic section can be viewed as a natural image, and the structures contained in the image are visible as areas of certain texture patterns. Then, these patterns can be described and analyzed by the image-based attributes.
There were indeed some studies that explored to some extent image-based texture attributes for seismic interpretation. For example, attributes based on the gray level co-occurrence matrix (GLCM) were applied to salt dome detection \cite[]{Gao2003,berthelot2013texture} and deep-marine facies discrimination \cite[]{gao2007application}; attributes derived from the Hilbert transform were utilized for seismic image segmentation \cite[]{pitas1992texture}; and Gabor filters were adopted for seismic image segmentation as well \cite[]{roster1998system}.

However, as successful as these applications were, we believe the exploration has been inadequate considering the tremendous success texture analysis has witnessed in the image processing community. Many advanced techniques developed in recent years have not been examined in the seismic context. More useful applications based on these attributes could be developed to fully exploit their potential. In particular, we believe texture attributes are suitable for serving as descriptors that characterize seismic volumes in terms of the various structures contained therein. Such descriptors are generic, capable of identifying the visible structures all at once, rather than being specific, targeted toward only a certain structure (e.g., salt dome or faults). Such generic descriptors are essential for computationally providing a comprehensive description of the subsurface environment, which can render an initial big picture with spots of possible interest highlighted to expedite the interpretation process.

Therefore, in this paper, we conduct a comparative study examining image-based texture attributes within the context of structure-based characterization of a seismic volume. To fit in the context of the application, we specifically focus on the spatial attributes that belong to the group of local descriptors. The local descriptors capture patterns of variations in visual elements such as intensities and edges in localized scales. The captured patterns are encoded into binary strings, accomplishing robust and computationally efficient texture representations. Such descriptors include the local binary pattern (LBP) \cite[]{ojala2002multiresolution}, the completed LBP (CLBP) \cite[]{guo2010completed}, the multi-scale CLBP (M-CLBP) \cite[]{guo2010completed}, the extended LBP (ELBP) \cite[]{Liu201286}, the completed local derivative pattern (CLDP) \cite[]{hu2016cldp}, and the local radius index (LRI) \cite[]{Zhai2013:LRI}. In addition to the above, we also include in our study two spatial attributes that are more familiar to the seismic interpretation community. One is the classic GLCM, and the other is the semblance, both of which have been used as texture attributes for seismic applications such as salt dome detection \cite[]{berthelot2013texture}.

\begin{figure}
    \centering
 	\includegraphics[width=0.7\linewidth]{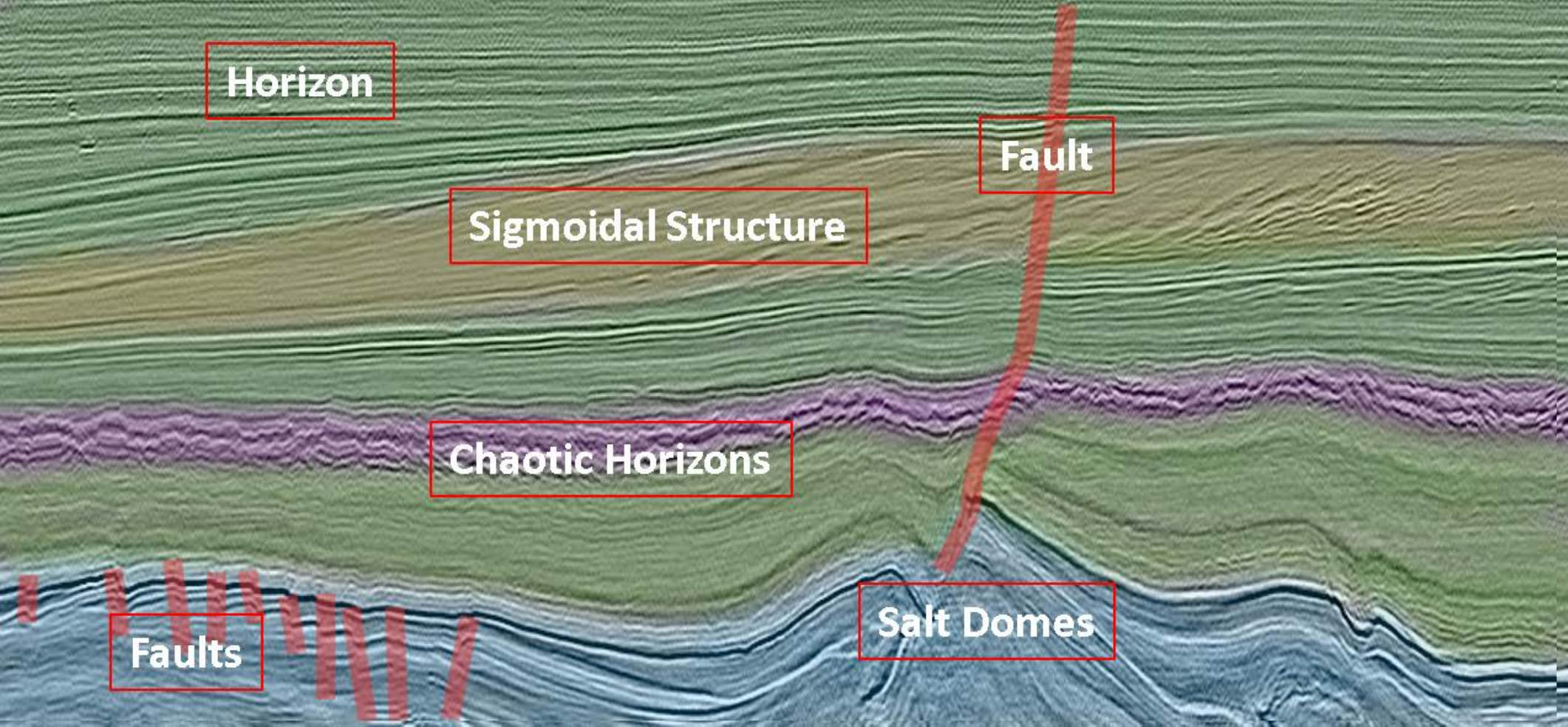}
 	\caption{An illustrative example of a manually labeled seismic section}
 	\label{fig:manualExample}
\end{figure}

To evaluate the performance of each attribute at characterizing subsurface structures, we adopt a newly-developed framework for interpretation, i.e., seismic volume labeling \cite[]{alaudah2016weakly}. With labeling, a data volume is automatically segmented into different structures and regions, each assigned with its corresponding label. The segmentation and the label assignment (or classification) are performed based on the extracted texture attributes. Thus, the labeling performance reflects the characterizing capability of the attributes. Figure~\ref{fig:manualExample} presents an illustrative example of a manually labeled seismic section. The labels highlight significant structures, which can help pinpoint spots of interest to interpreters to make their exploration more focused and effective. It is worth noting that, the purpose of the labeling is not to accurately delineate the subsurface structures in a given seismic volume. Rather, the labeling is intended to generate an initial map that highlights approximate locations of the structures so that an interpreter can quickly determine where to look further. The labeling simultaneously identifies all structures visible in the data. Techniques customized for a specific structure can then be applied at the will of the interpreter for more focused and refined examination.

The rest of the paper is organized as follows. First, we discuss the various attributes of interest. Then, we introduce the computational framework we use for seismic volume labeling. The attributes will be examined in the labeling experiments afterwards. Finally, we draw conclusions according to our observations in the experiments. The related source code will be available on our website (https://ghassanalregib.com/). 

%% file: glcm.tex
The GLCM-based attributes have been widely accepted as useful tools for texture analysis since they were proposed four decades ago \cite[]{Haralick1973}. The GLCM is a matrix that describes the co-occurrence pattern between gray levels of two neighboring pixels along a certain direction in an image. In essence, it represents a two-dimensional histogram that approximates the joint probability distribution of the neighboring gray values. It can capture textural patterns for the selected neighborhood along the prescribed direction. For example, high values away from the diagonal in a GLCM reveal sharp changes in gray level, whereas high values close to the diagonal indicate small variations.

Given an 8-bit gray scale image $\boldsymbol I$ of dimension $M \times N$, where $I[m,n]$ represents the gray value at location $[m,n]$. The GLCM (i.e., the co-occurrence matrix of gray values), $\boldsymbol C$, is calculated as below:
\begin{equation}
C_{d,\theta}[i,j] = \sum_{m=1}^M \sum_{n=1}^N \delta[m,n],
\label{eq:GLCM}
\end{equation}
where
\begin{equation}
\delta[m,n] = \begin{cases}1, & \text{if } I[m,n]=i \text{ and } I[m+\Delta m,n+\Delta n]=j \\ 0, & \text{otherwise} \\
\end{cases},
\end{equation}
$d = \sqrt{\Delta m^2+\Delta n^2}$, $\theta = tan^{-1}\left(\frac{\Delta n} {\Delta m}\right)$, and $C_{d,\theta}[i,j]$ represents the number of occurrences of gray level $j$ adjacent to gray-level $i$ separated by a distance $d$ in direction $\theta$. Based on the GLCM, assuming the total number of gray levels is $K$, the corresponding probability mass function $\boldsymbol P$ is computed as follows:
\begin{equation}
P_{d,\theta}[i,j]= \frac{C_{d,\theta}[i,j]}{\sum_{s=1}^K \sum_{t=1}^K C_{d,\theta}[s,t]}.
\label{eq:NormalizedGLCM}
\end{equation}

Once $\boldsymbol P$ is available, various GLCM-based attributes can be generated. For simplicity, we use $P_{ij}$ to represent $P_{d,\theta}[i,j]$, and list below the typical attributes:
\begin{equation}
\mathrm{Contrast}=\sum_i\sum_j\left|i-j\right|^2P_{ij},
\end{equation}
\begin{equation}
\mathrm{Entropy}=-\sum_i\sum_jP_{ij} \log P_{ij},
\end{equation}
\begin{equation}
\mathrm{Energy}=\left[\sum_i\sum_jP_{ij}^{2}\right]^{\frac{1}{2}},
\end{equation}
\begin{equation}
\mathrm{Homogeneity} = \sum_i \sum_j \frac{1}{1+{(i-j)}^2}P_{ij},
\end{equation}
\begin{equation}
\mathrm{Mutual~Information} = \sum_i \sum_j p_{ij} \log \frac{p_{ij}}{p_i p_j}, \text{ where } p_i = \sum_j p_{ij} \text{ and } p_j = \sum_i p_{ij}.
\label{eq:MutualInformation}
\end{equation}

Among these attributes, the GLCM contrast is a measure of the local gray-level variations; the GLCM entropy describes the spatial disorder or complexity in textures; and the GLCM energy measures the pixel pair repetitions, also called texture uniformity or angular second moment. Their values are low for smooth regions, and high for areas with rich texture. On the contrary, both the GLCM homogeneity and the GLCM mutual information \cite[]{beghdadi2015} will show high values for smooth areas and low values for complex textures, with the former being inversely correlated to the GLCM contrast and the latter presenting the dependency among the neighbors.

%% file: semblance.tex
The semblance attribute describes the similarity in a certain spatial neighborhood. Although it is not commonly used for texture image analysis, it has been used for seismic interpretation applications such as detection of faults \cite[]{wang2017interactive} and salt domes \cite[]{berthelot2013texture}, both of which are structures of interest in this study. In this paper, we adopt the well-known dip-guided semblance developed by~\cite{marfurt1998}.

%% file: lbp.tex
The local binary pattern (LBP) is a simple and efficient texture attribute, which has become a standard local texture descriptor in the spatial domain \cite[]{ojala2002multiresolution}. It describes the intensity difference between a pixel and its local circular neighborhood, denoted by $(P, R)$, where $P$ defines the number of pixels evenly distributed on the circular neighborhood with radius $R$. To ensure robustness against intensity changes, LBP employs the signs of the differences instead of the exact values to form unique binary codes for the description of local texture patterns.

LBP is calculated as follows:
\begin{equation}
\label{equ:lbp}
LBP_{P,R}=\sum\limits_{p=0}^{P-1}s(g_p-g_c)\cdot 2^p,\quad
s(x)=
\left\{
\begin{aligned}
1,&\mbox{ if } x\geq 0\\
0,&\mbox{ otherwise}
\end{aligned},
\right.
\end{equation}
where $g_c$ and $g_p$, $p=0,1,\cdots,P-1,$ represent the intensity of the center pixel and its corresponding neighboring pixels, respectively. Function $s(\cdot)$ extracts the sign information of the differences, with the value being $1$ for non-negative ones and $0$ for negative ones. As the equation shows, LBP encodes the local intensity variation (i.e., the intensity difference between a pixel and its neighbors) into a binary code, resulting in a computationally efficient representation.

The $LBP_{P,R}$ calculation will produce $2^P$ binary patterns. However, a rotation of a texture pattern may lead to different coding results, because $g_0$ is always assigned to an element in a fixed location (e.g., the one to the right of $g_c$). To account for the rotation effect, binary patterns with the same circularly shifted code are grouped into one rotation invariant pattern, denoted by $LBP_{P,R}^{ri}$. This grouping reduces the total number of possible patterns. As an example, for $P=8$, the $LBP_{P,R}^{ri}$ scheme can reduce the total number of patterns from $2^8=256$ to $36$.

In addition, based on the fact that some binary patterns have higher frequencies of occurrence in texture images than others, \cite{ojala2002multiresolution} defined uniform patterns that contain at most two bitwise transitions (i.e., $0$ to $1$ or $1$ to $0$) when traversed circularly. Such uniform patterns are determined as follows:
\begin{equation}
\label{equ:lbp_u2}
LBP_{P,R}^{riu2}=
\left\{
\begin{aligned}
&\sum\limits_{p=0}^{P-1}s(g_p-g_c),\mbox{ if }U\left(LBP_{P,R}^{ri}\right)\leq2\\
&P+1,\quad\ \ \qquad\mbox{ otherwise}
\end{aligned},
\right.
\end{equation}
where function $U(LBP_{P,R}^{ri})$ counts the bitwise transitions for each rotation invariant pattern, and superscript ``$riu2$" indicates ``uniform patterns with rotation invariance." Introducing the uniform patterns further reduces the number of patterns. When $P=8$, the number of $LBP_{P,R}^{riu2}$ patterns changes from $36$ (for $LBP_{P,R}^{ri}$) to $10$.

%% file: lbp_variants.tex
Although LBP is simple and efficient for texture analysis, its performance can be further enhanced by including more local information in addition to the neighbor-center sign information. Following this strategy, a series of LBP variants have been developed. In this study, we select several typical examples among them, including CLBP \cite[]{guo2010completed}, M-CLBP \cite[]{guo2010completed}, ELBP \cite[]{Liu201286}, and CLDP \cite[]{hu2016cldp}. As summarized in Table \ref{table_lbp_variants}, these techniques vary in two aspects: 1) what local information to include in the descriptive components; and 2) how to incorporate such information. To obtain uniform patterns with rotation invariance, the same ``$riu2$'' mapping as adopted for LBP is used.

\begin{figure}[tbp]
\begin{centering}
\footnotesize
\begin{tabular}{c c}		
    \includegraphics[width=0.25\linewidth]{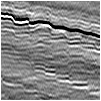}
	& \includegraphics[width=0.25\linewidth]{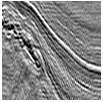} \\
	\includegraphics[width=0.45\linewidth]{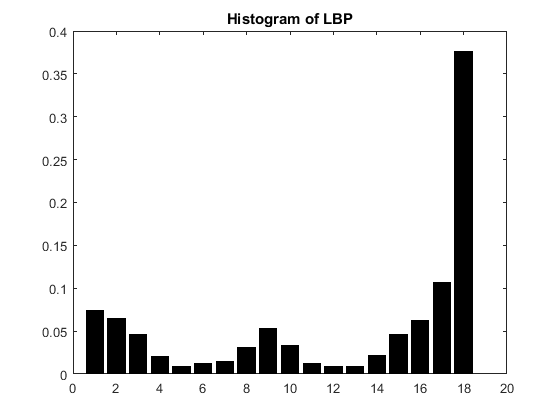}
    & \includegraphics[width=0.45\linewidth]{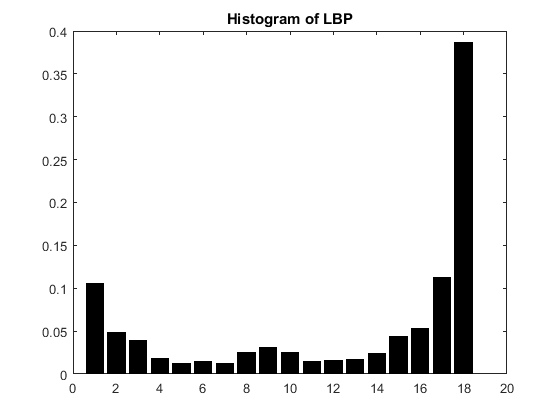} \\
    \includegraphics[width=0.45\linewidth]{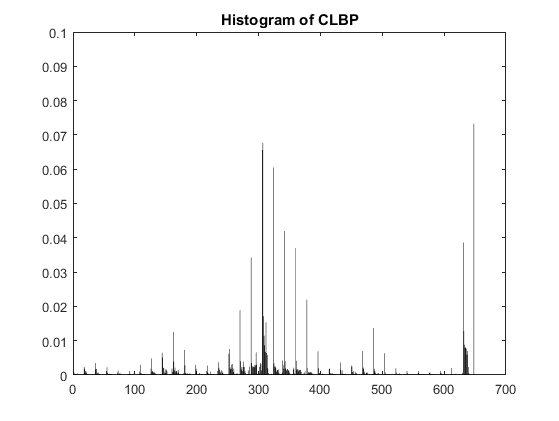}
	& \includegraphics[width=0.45\linewidth]{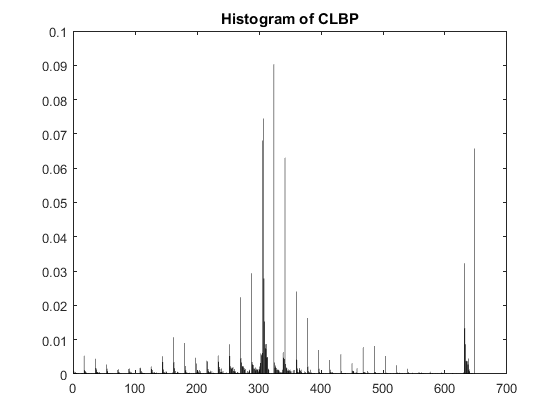} \\
	\includegraphics[width=0.45\linewidth]{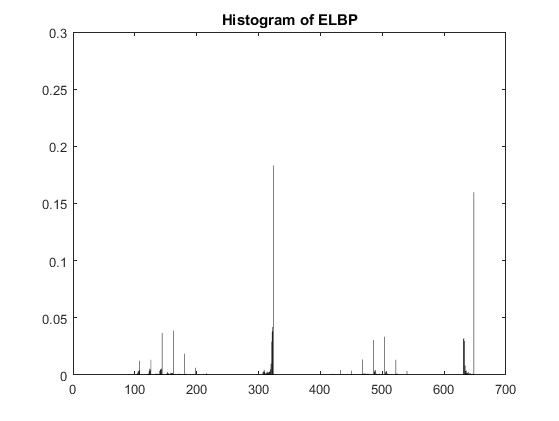}
    & \includegraphics[width=0.45\linewidth]{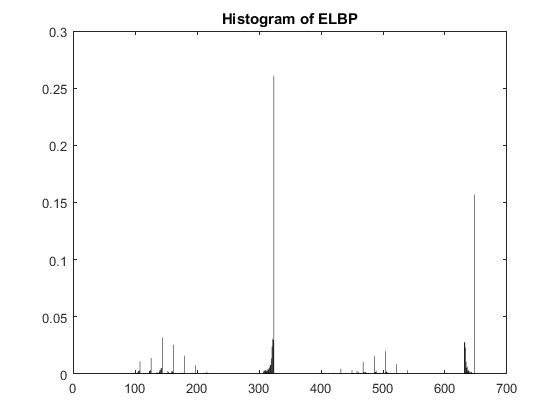} \\
\end{tabular}
\caption{Example histograms calculated using LBP and some variants for typical structures. The left column is for faults, and the right column is for salt dome. The histograms are ordered as (top to bottom): LBP, CLBP, and ELBP.}
\label{fig:example_attributes}
\end{centering}
\end{figure}

For texture analysis, usually the local descriptors are not examined directly. Instead, histograms are generated from the attributes and used by the algorithms for analysis. In other words, it is the probability distribution of the attributes that helps provide a robust representation of texture patterns. When there are a few descriptive components involved, they are combined to yield either a joint histogram or a concatenated histogram. In Figure~\ref{fig:example_attributes}, histograms obtained for some example subsurface structures using LBP and its variants are given for an illustration.

\begin{table} [tbp]
\footnotesize
\centering
\caption{Comparison of LBP and Variants}
\label{table_lbp_variants}
\begin{tabular}{|c|c|c|}
    \hline
    \textbf{Techniques} & \textbf{Components} & \textbf{Notes}\\
    \hline\hline
    LBP & Neighbor-center Difference & The original form \\
    \hline
    CLBP & \begin{tabular}[c]{@{}c@{}} Neighbor-center Difference ($CLBP\_S$) \\ Neighbor Intensity ($CLBP\_M$)\\ Center Intensity ($CLBP\_C$) \end{tabular} & \begin{tabular}[c]{@{}c@{}} $CLBP\_S$ is identical to the orignal LBP; \\ intensity of neighbors and center also added \end{tabular} \\
    \hline
    M-CLBP & \begin{tabular}[c]{@{}c@{}} Combination of CLBP at different scales \\ (i.e., different values for the radius $R$) \end{tabular} & \begin{tabular}[c]{@{}c@{}} Multiscale implementation of CLBP \\ for a more comprehensive characterization \end{tabular} \\
    \hline
    ELBP & \begin{tabular}[c]{@{}c@{}} Neighbor Intensity ($ELBP\_NI$) \\ Center Intensity ($ELBP\_CI$) \\ Radial Difference ($ELBP\_RD$) \end{tabular} & \begin{tabular}[c]{@{}c@{}} Ignores the neighbor-center difference; \\ $ELBP\_RD$ incorporates cross-scale correlation, \\ while M-CLBP considers each scale separately \end{tabular} \\
    \hline
    CLDP & \begin{tabular}[c]{@{}c@{}} Neighbor-center Difference ($CLDP\_S$) \\ Neighbor Intensity ($CLDP\_M$) \\ Center Intensity ($CLDP\_C$) \\ Radial Sign Difference ($CLDP\_D$) \end{tabular} & \begin{tabular}[c]{@{}c@{}} Adds to CLBP cross-scale correlation \\ ($CLDP\_D$), which is different from $ELBP\_RD$ \end{tabular} \\
    \hline
\end{tabular}
\end{table}

%% file: lri.tex
LBP and its variants, as discussed above, all examine the local variation of the pixel intensities. For texture images, edges are also commonly observed. Although LBP-like attributes implicitly capture the edge information, a direct description of the spatial distribution of edges can be more effective. For this purpose, the local radius index (LRI) was proposed by \cite{Zhai2013:LRI}, which characterizes texture patterns using the local distribution of distances between adjacent edges along a particular angle.

Based on how the local index is computed, there are two variants of LRI: LRI-A and LRI-D. For LRI-A, the inter-edge distance in a given direction is calculated, which represents the width of adjacent smooth regions. In contrast, for LRI-D, the distance is measured from a pixel to its nearest edge, i.e., the boundary of the next smooth region. In this paper, we will only discuss LRI-A, because we did not observe any significant difference in performance between the two in our experiments. An example is shown in Figure~\ref{fig:LRI} illustrating how to compute LRI-A for a small group of pixels. The corresponding procedure is given below.

For pixel $x_i$ and direction $d = 1,\cdots,8$, let $a_{dj}$ denote $j$ neighboring pixels in direction $d$, where $j=1, \cdots,K$, then
\begin{equation}
\label{eq:LRI-A}
LRIA_d=\begin{cases}
0, & \text{if } |x_i-a_{d1}| \le  T\\
\text{min }(j,K), & \text{if } a_{dj} > x_i+T \text{ for } j=1,\cdots,K \text{ but not K+1 }\\
\text{max}(-j,-K), & \text{if } a_{dj} < x_i-T \text{ for } j=1,\cdots,K \text{ but not K+1 }\end{cases},
\end{equation}
where we use threshold $T$ to define an edge, and $K$ to prescribe the size of the texture elements. Consequently, $T$ controls the noise sensitivity, while $K$ determines the computational complexity. In the volume labeling experiments to be discussed in the next two sections, we used $T=0.5\sigma$ and $K=3$, where $\sigma$ stands for the standard deviation of the local intensities.

\begin{figure}
\centering
\includegraphics[scale=0.4]{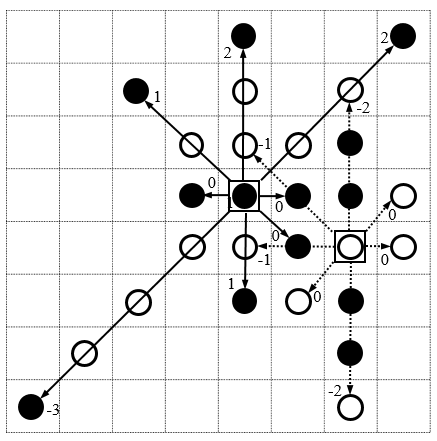}
\caption{An example illustrating how to compute LRI-A, where black dots refer to edge pixels and white dots are non-edge pixels.}
\label{fig:LRI}
\end{figure}

%% file: labeling.tex
\begin{figure}[t]
    \centering
 	\includegraphics[width=\linewidth]{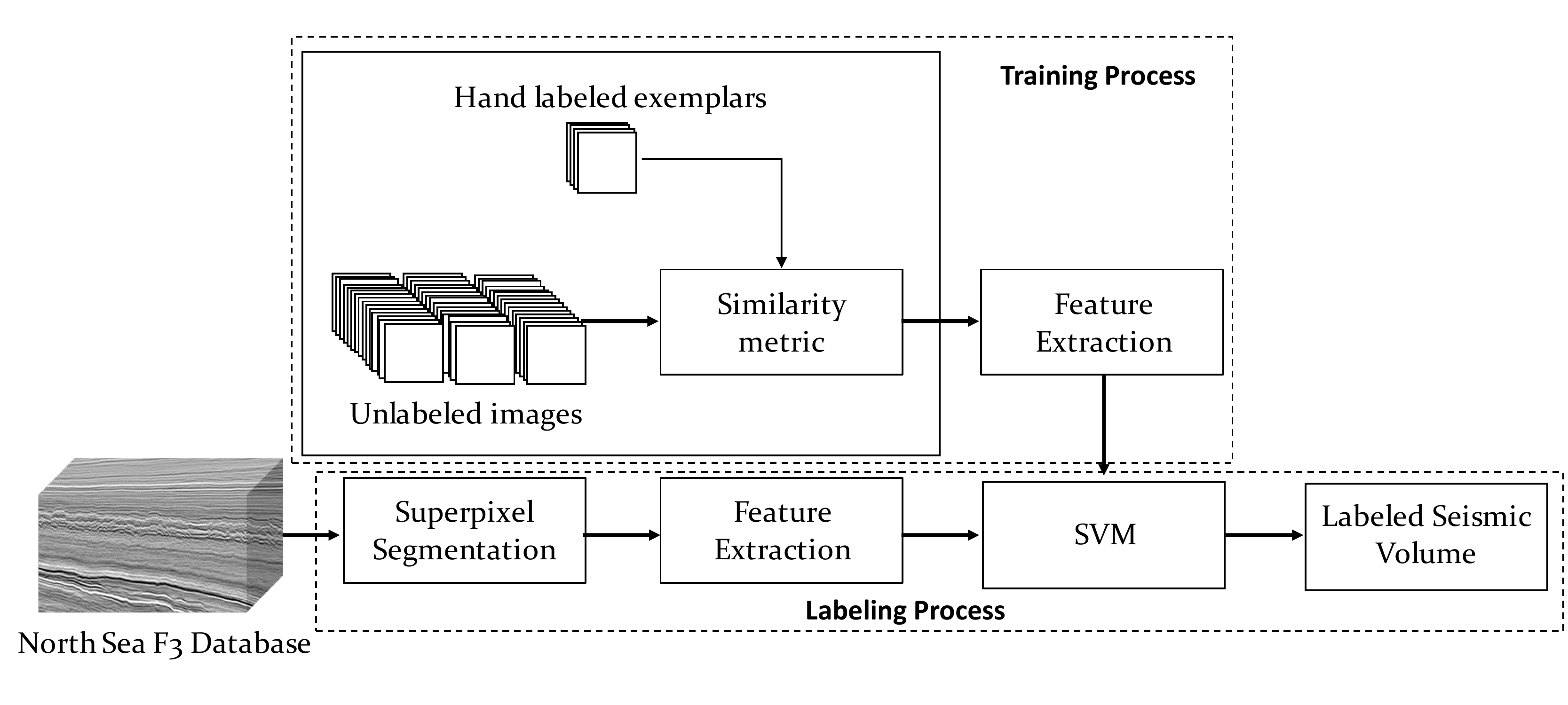}
 	\caption{An illustration of the framework for computational seismic volume labeling. \cite[]{alaudah2016weakly}}
 	\label{fig:Main_Diagram}
\end{figure}

To evaluate the capability of the texture attributes for characterizing subsurface structures, we adopt our newly-developed framework for interpretation (i.e., computational seismic volume labeling) \cite[]{alaudah2016weakly}. The objective is to automatically divide a data volume into segments consisting of various structures and assign each structure with its corresponding label. As demonstrated in Figure~\ref{fig:Main_Diagram}, this task is accomplished in a setting that combines segmentation, retrieval, and  supervised classification. It consists of a training process and a labeling (or testing) process.
 
\begin{figure}
\begin{center}
\begin{tabular}{c c c c}
\includegraphics[width=0.2\linewidth]{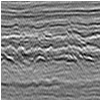} & \includegraphics[width=0.2\linewidth]{Fig/F.png} &
\includegraphics[width=0.2\linewidth]{Fig/S.png} & \includegraphics[width=0.2\linewidth]{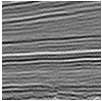}\\
(a) & (b) & (c) & (d)\\
\end{tabular}
\end{center}
\caption
{Sample image patches (patch size: $99 \times 99$) from the four defined classes of structures \cite[]{alaudah2016weakly}: (a) Chaotic; (b) Faults; (c) Salt dome; (d) Other.}
\label{fig:images}
\end{figure}

Our training process involves three steps. The first step is data extraction, in which, given a small set of manually labeled exemplar patches, a large number of image patches with the same geological structures are automatically extracted to form the training samples. To do this, we use seismic sections extracted along the crossline direction of the widely used Netherlands North Sea Offshore F3 Block \cite[]{database}. All sections are normalized to remove contrast and mean variations between sections. Patches containing various subsurface structures are extracted from seismic sections. We define three structures that are of interest, namely, \texttt{Chaotic} layers, \texttt{Faults}, and \texttt{Salt dome}. We also define the \texttt{Other} class for patches that does not contain any of the previous three structures. Examplar patches from each class are shown in Figure~\ref{fig:images}. Here, the patches are set to a fixed size $99 \times 99$ to make sure the representative texture patterns are captured for each structure of interest, because the labeling is designed to simultaneously identify all structures, not a specific structure. For the same reason, we can only use a square patch, even though rectangular ones may work better for directional structures such as faults.

\begin{figure}[!h]
    \centering
 	\includegraphics[width=0.8\linewidth]{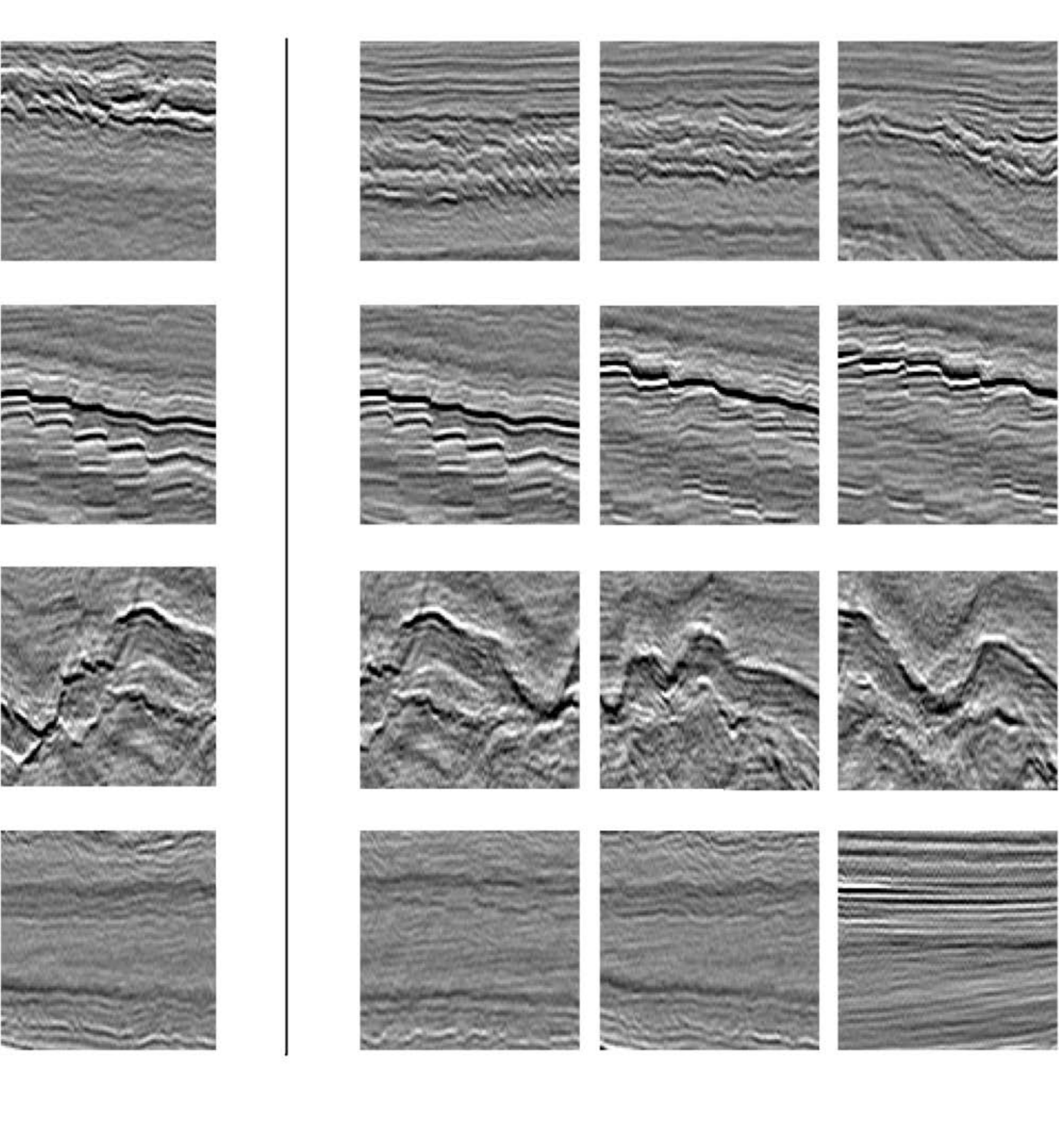}
 	\caption{Illustrative examples of automatically extracted image patches. Here the first column shows four exemplar patches; the second column shows automatically extracted patches, where the similarity scores are among the top 10 of all 500 similarity scores for each exemplar patch; the third column represents extracted patches whose similarity scores are ranked from 11 to 100; and for the fourth column, the similarity scores are ranked from 101 to 500. }
 	\label{fig:examplePatch}
\end{figure}

We extract six exemplar patches (two from \texttt{Salt dome}, two from \texttt{Other}, and one from each of the rest) and manually label them. We then automatically extract 500 image patches from the seismic sections for each type of the defined structures based on their similarity to the exemplar patches, measured according to a recently proposed texture similarity metric \cite[]{dtsim}. These extracted patches form training samples for a succeeding classification step. The automatic similarity-based patch extraction assists in avoiding the time-consuming manual labeling of the training dataset. Figure~\ref{fig:examplePatch} presents some examples of the automatically extracted patches, which show excellent consistency comparing to their respective exemplar patches.

The second step of the training process is to extract texture attributes from each patch in the training set. Before applying any of the techniques discussed in this paper to do the attribute extraction from an image patch $X_i$, we first calculate the Hadamard product of this patch with a two-dimensional Gaussian kernel of the same size, $G$. This kernel gives more weights to the structures at the center of the patch and less to those on the periphery, thus emphasizing local spatial correlations in seismic data. The procedure can be expressed as follow:
\begin{equation}
	\widetilde{X}_i = X_i \odot G,
\end{equation}
where $\odot$ is the Hadamard product, and the Gaussian kernel is defined as
\begin{equation}
	G[x,y] = e^{ \frac{(x-\mu_x)^2 + (y - \mu_y)^2}{2\sigma^2} }, 	
\end{equation}
where $\mu_x$ and $\mu_y$ are the $x$- and $y$- coordinates of the center of $X_i$, respectively. The value of $\sigma$ was set to 25 in our experiments so that pixels in the corners of the patch have weights of less than $1\%$.

After this pre-processing, one of the eight techniques is applied to the patch to generate texture attributes. For LBP, the direct output generated for a given patch is a map of LBP codes of the same size as the original patch. For texture classification, a histogram is typically calculated from the LBP map and fed into the classifier as the final attributes. For other local descriptors (e.g., CLBP, M-CLBP, ELBP, CLDP, and LRI), multiple components are calculated in the form of multiple attribute maps (e.g., maps for CLBP-S, CLBP-C, and CLBP-M, respectively). In such cases, either a joint histogram or a concatenation of the individual histograms are formed for the classifier to use. To retain consistency, the semblance attribute is converted to a histogram when examined for the labeling task. The only exception is for the GLCM attributes, each of which is already computed from a two-dimensional histogram (i.e., the co-occurrence matrix) based on the patch. In this case, we combine different GLCM attributes into one vector as an input to the classifier.

In the third step of the training process, histograms generated from the texture attributes are used to train the classifier to establish statistical models for each category of subsurface structures of interest. In this research, we choose the support vector machine (SVM) \cite[]{vapnik} as the classifier, which is a powerful binary classification algorithm. It seeks to find the optimal separating hyperplane between two classes by identifying the one with the maximum margin. Since we have a multi-class classification problem, we train four hard-margin SVMs with linear kernels using the one-versus-all (OVA) approach.

Once the training process is complete, the label assignment is fulfilled in the labeling process, which is divided into three steps. First, a segmentation is performed in which each seismic section to be labeled is automatically divided into segments according to structures. To accomplish this purpose, we employ a superpixel-based segmentation approach that groups neighboring image pixels of similar appearance into a cluster or a superpixel. Each superpixel is treated as a single unit in the following processing. In this work, we adopt the superpixel segmentation algorithm based on the simple linear iterative clustering (SLIC) \cite[]{SLIC}. In the original SLIC, vectors in the form of $[l, a, b, x, y]$ are generated for each pixel in an image to be segmented, where $l$, $a$, and $b$ are the three components of the $Lab$ color model, and $x$, $y$ are the coordinates for each pixel. Then clustering is performed in a space formed by these vectors to obtain the superpixels. Because seismic images are in grayscale, we compute vectors for the pixels in a modified form, i.e., $[l, g_x, g_y, x, y]$, in which $g_x$ and $g_y$ refer to the gradient along the x- and y- directions, respectively. Afterwards, we generate superpixels by clustering the vectors.

In the second step of labeling, similar to the training process, texture attributes are extracted for each segment or superpixel. Typically, the size of a superpixel is smaller than that of the image patches in the training dataset. To make sure that the attribute extraction is consistent between the training and the labeling processes, in this work, we select a neighborhood centered around the centroid of the superpixel, which is of the same size as that of the training patches. Attributes are extracted from this selected neighborhood (rather than the smaller area covered by the superpixel) to represent the superpixel. Then, in the last step of labeling, histograms of attributes generated for each superpixel are fed into the SVM classifier. They are compared against the trained SVM models to determine which of the four types of structures each superpixel belongs to. Thus, each one of them is assigned with an appropriate label.

The training and labeling processes consist of four components: data extraction (for training), attributes extraction (for both training and labeling), segmentation (for labeling), and classification (for both training and labeling). Training the classifier is considered as part of the classification component.

%% file: results.tex
We examined texture attributes within the framework of the computational labeling. As mentioned above, we used seismic sections extracted from the Netherlands North Sea Offshore F3 Block \cite[]{database}. We specified four structures to be labeled: \texttt{Chaotic} layers, \texttt{Faults}, \texttt{Salt dome}, and \texttt{Other}. After training the SVM classifier as described in the previous section, four selected seismic sections, such as crossline 61, 211, 231, and 281 were labeled using each attribute. In Table~\ref{table_settings}, we list parameter settings we used to generate each attribute during the experiments.

\begin{table} [bp]
\footnotesize
\centering
\caption{Parameter settings used for each attribute}
\label{table_settings}
\begin{tabular}{|c|c|}
    \hline
    \textbf{Attribute} & \textbf{Settings}\\
    \hline \hline
    GLCM & Combined: Contrast, Entropy, Energy, Homogeneity, Correlation, and Mutual Information\\
    \hline
    Semblance & Neighborhood: $3 \times 3$\\
    \hline
    LBP & Radius ($R$): $2$; Samples ($P$): $16$; Mapping: $riu2$\\
    \hline
    CLBP & Radius ($R$): $2$; Samples ($P$): $16$; Mapping: $riu2$\\
    \hline
    M-CLBP & Radius ($R$): $1, 2, 3$; Samples ($P$): $8, 16, 24$; Mapping: $riu2$\\
    \hline
    ELBP & Radius ($R$): $2$; Samples ($P$): $16$; Mapping: $riu2$\\
    \hline
    CLDP & Radius ($R$): $2$; Samples ($P$): $16$; Mapping: $riu2$\\
    \hline
    LRI & LRI-A; Threshold for edge ($T$): $\sigma/2$; Threshold for size ($K$): $3$; Directions: $8$\\
    \hline
\end{tabular}
\end{table}

The labeling performance was evaluated both subjectively by visual inspection and objectively in terms of four metrics commonly accepted in the semantic segmentation literature \cite[]{long2015fully}: pixel accuracy (PA), mean class accuracy (MCA), mean intersection over union (MIU), and frequency-weighted intersection over union (FWIU). The four metrics all range from $0$ to $1$, with a greater value indicating a better performance. PA counts the overall rate of correctly labeled pixels, and MCA calculates the average rate of such labels among all classes. However, both of them neglect wrongly labeled pixels that have a negative impact on labeling performance. To account for such false labels, MIU and FWIU can be employed. Assuming that manually labeled data from an expert interpreter is available as the ground truth, the metrics are defined as below:
\begin{equation}
	PA = \frac{\sum_i n_{ii}}{\sum_i t_i},
\end{equation}
\begin{equation}
	MCA = \frac{1}{n_c} \sum_i \frac{n_{ii}}{t_i},
\end{equation}
\begin{equation}
	MIU = \frac{1}{n_c} \sum_i \frac{n_{ii}}{t_i + \sum_j n_{ji} - n_{ii}},
\end{equation}
\begin{equation}
	FWIU = \frac{1}{\sum_k t_k} \sum_i \frac{t_i n_{ii}}{t_i + \sum_j n_{ji} - n_{ii}},
\end{equation}
where $n_{ji}$ refers to the number of pixels that belong to class $y_j$ but misclassified as $y_i$, $n_c$ indicates the number of classes, and $t_i$ denotes the total number of pixels in class $y_i$.

\begin{figure}[tbp]
\begin{centering}
\footnotesize
\begin{tabular}{c c}		
    \includegraphics[width=0.334\linewidth]{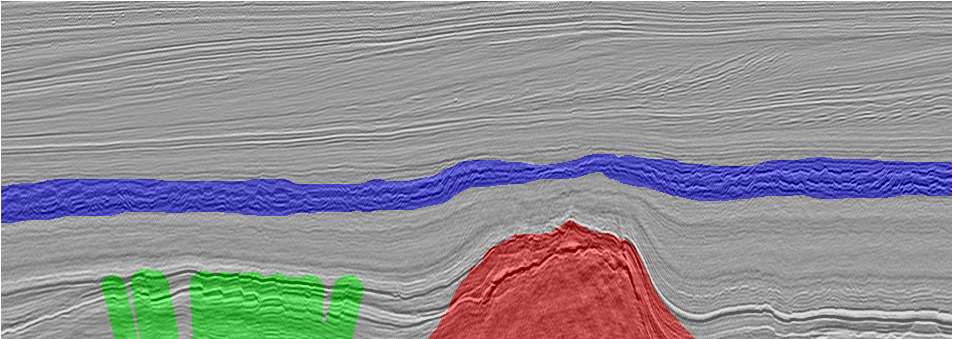} & \\
    (a) & \\
    \includegraphics[width=0.4\linewidth]{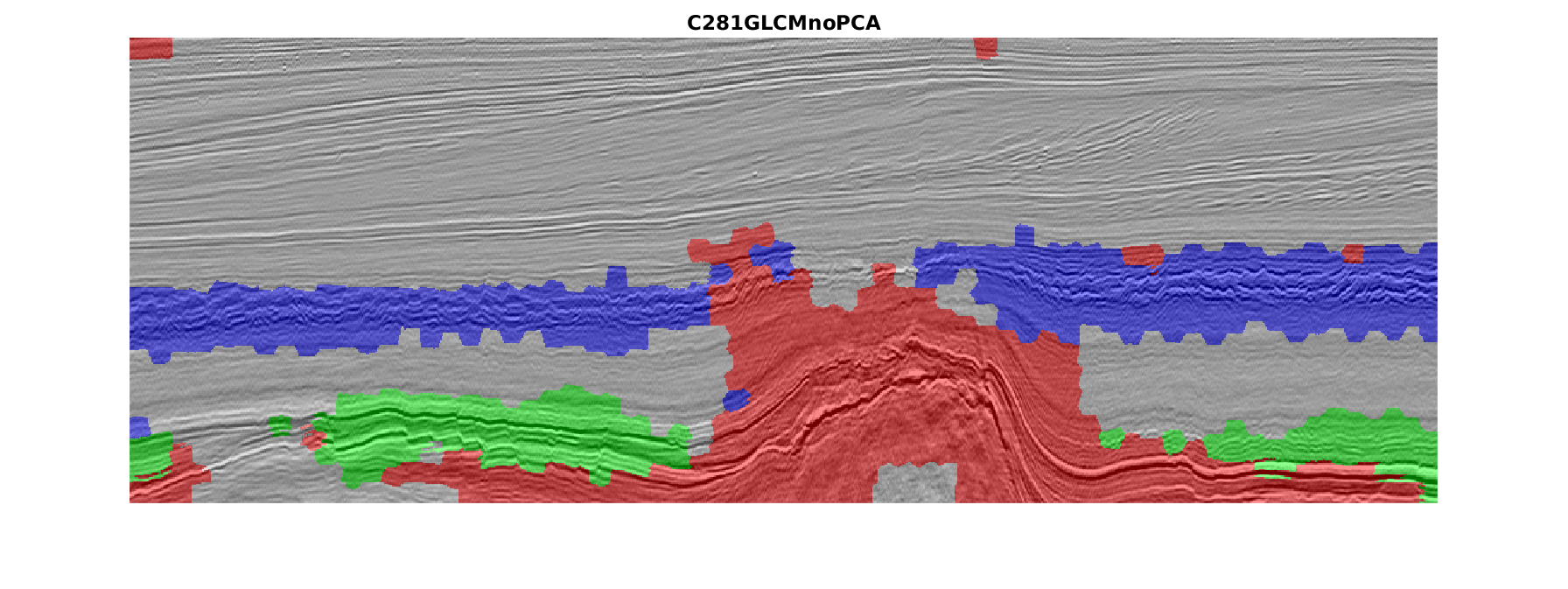} &  \includegraphics[width=0.4\linewidth]{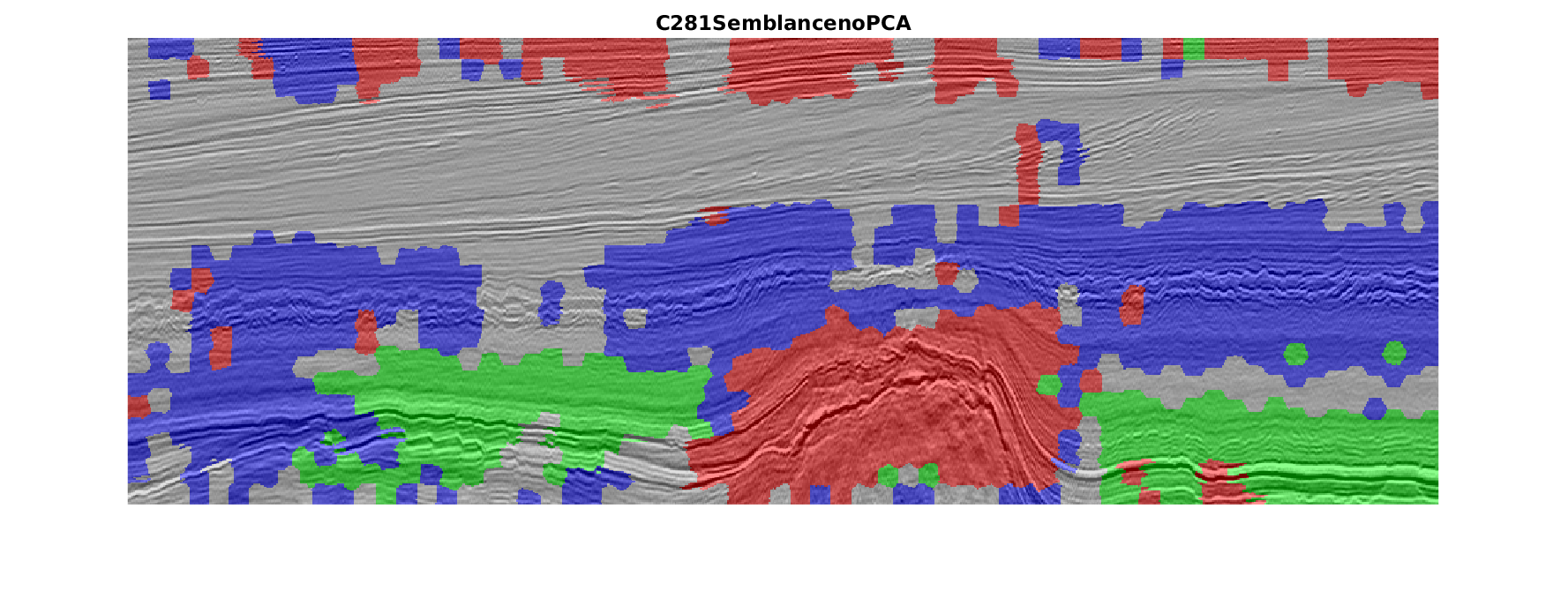} \\
    (b) & (c) \\
	\includegraphics[width=0.4\linewidth]{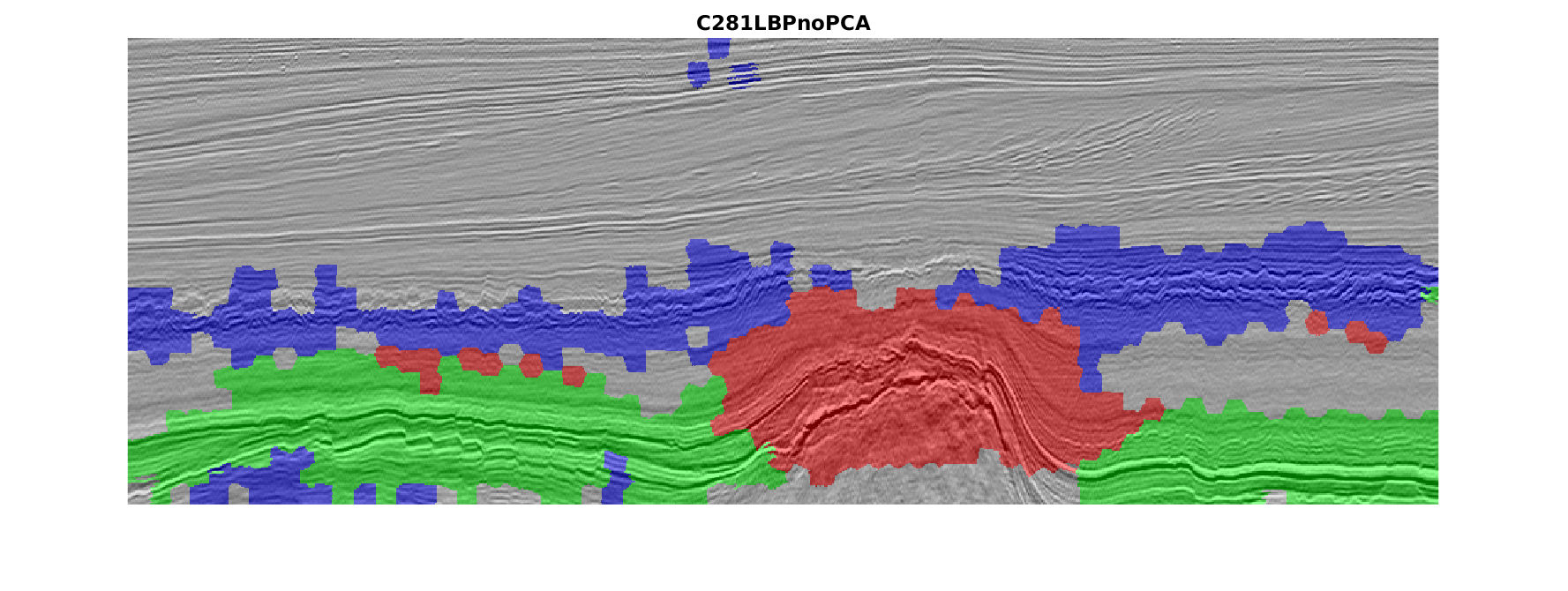} & \includegraphics[width=0.4\linewidth]{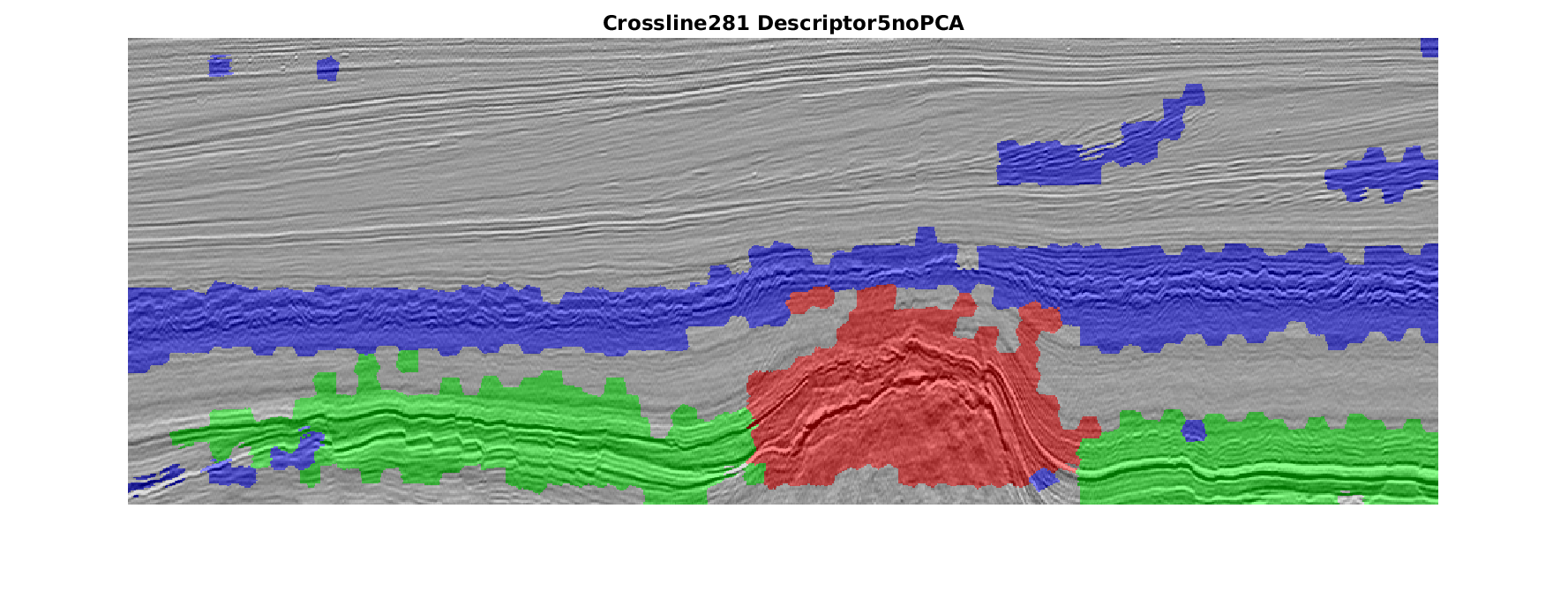} \\
    (d) & (e) \\
	\includegraphics[width=0.4\linewidth]{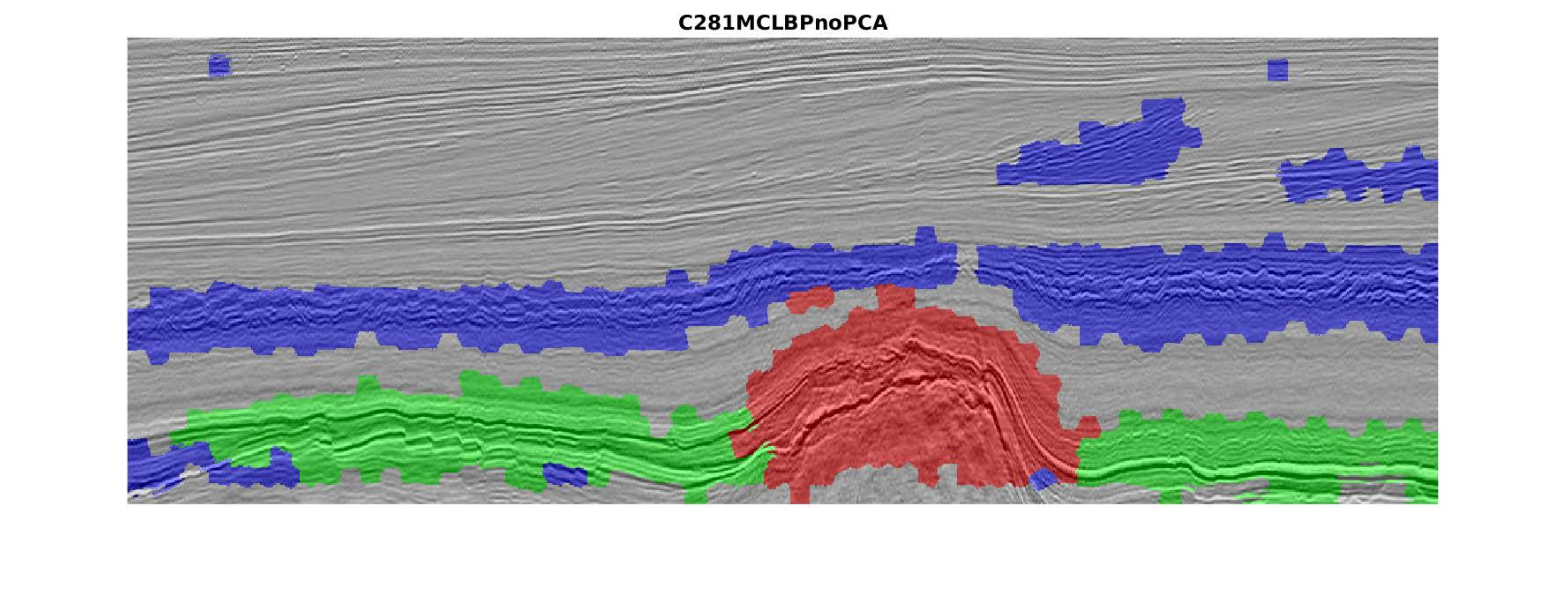} &
	\includegraphics[width=0.4\linewidth]{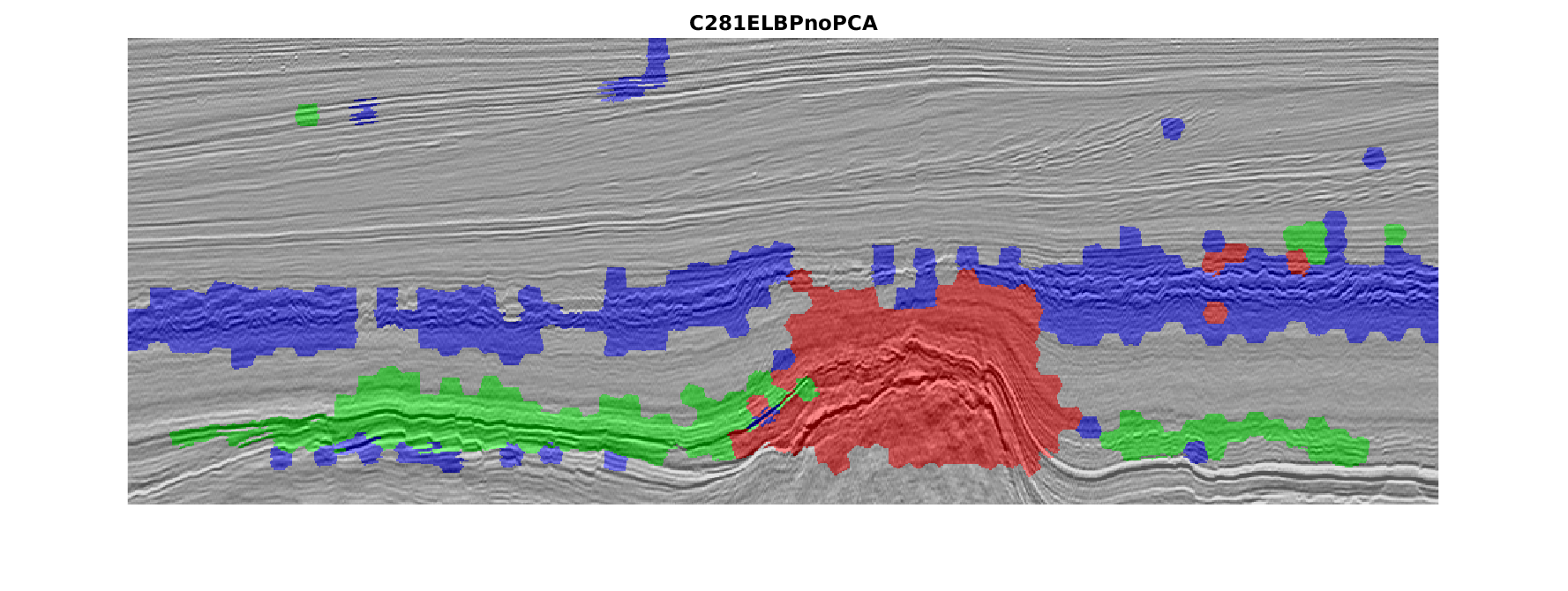} \\
    (f) & (g) \\
	\includegraphics[width=0.4\linewidth]{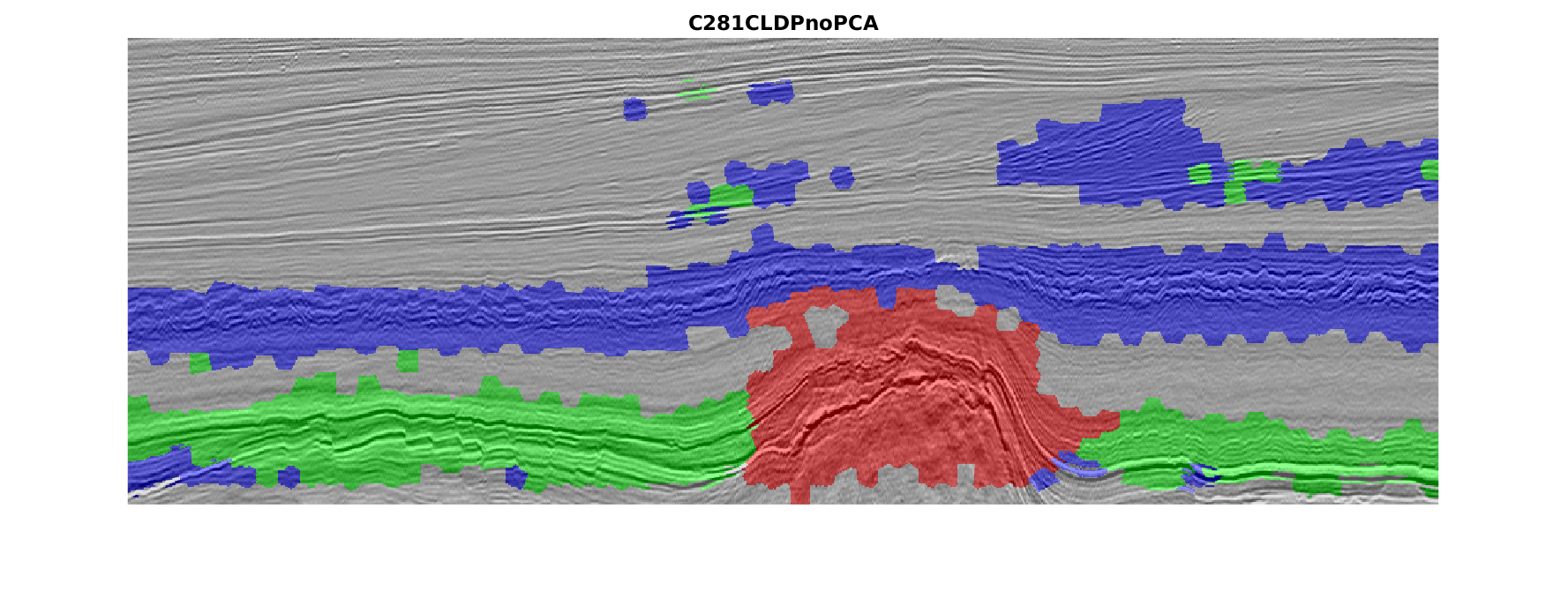} & 
	\includegraphics[width=0.4\linewidth]{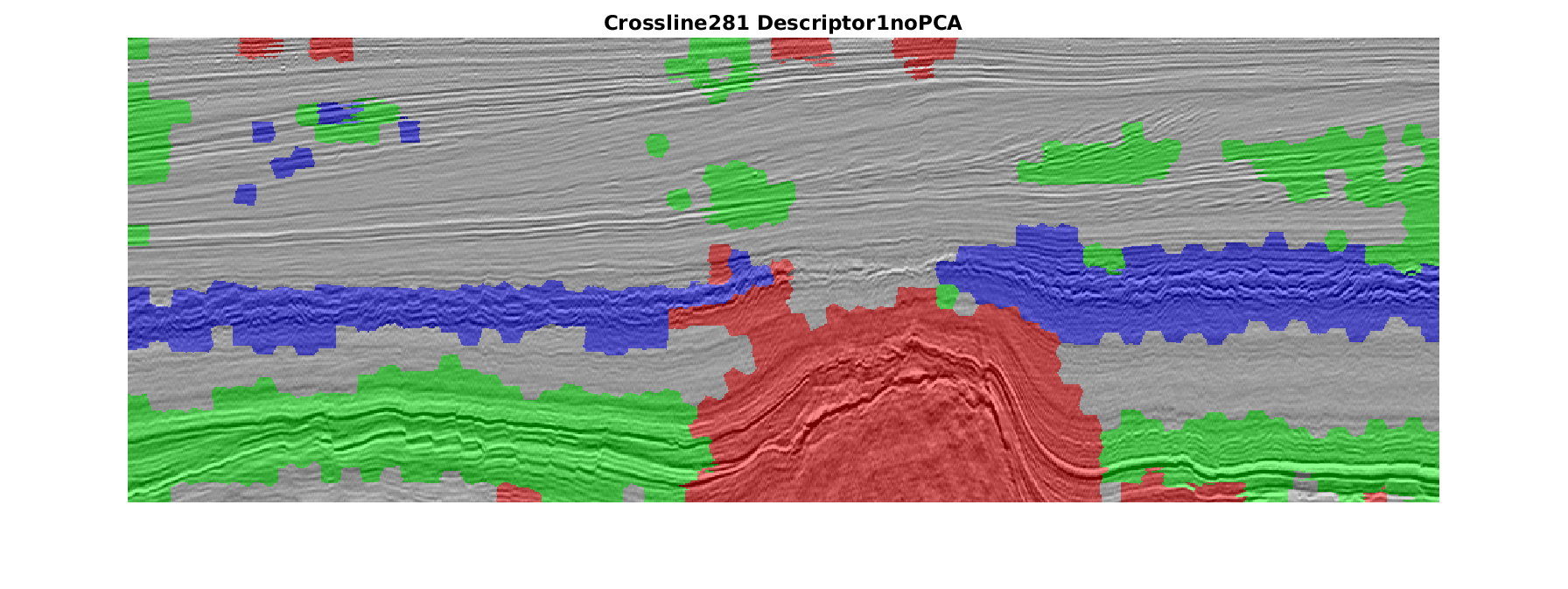} \\
    (h) & (i) \\
\end{tabular}
\caption{Labeling results for the seismic section at crossline 281, in the Netherlands North Sea F3 block database \cite[]{database}, using different texture attributes. The \texttt{Chaotic} class is in blue, \texttt{Faults} is in green, \texttt{Salt Dome} is red, and \texttt{Other} is grey. (a) Manual (manually labeled result overlaid upon the original data); (b) GLCM; (c) Semblance; (d) LBP; (e) CLBP; (f) M-CLBP; (g) ELBP; (h) CLDP; (i) LRI.}
\label{fig:results}
\end{centering}
\end{figure}

\begin{figure}[tbp]
\begin{centering}
\footnotesize
\begin{tabular}{c c}		
	\includegraphics[width=0.334\linewidth]{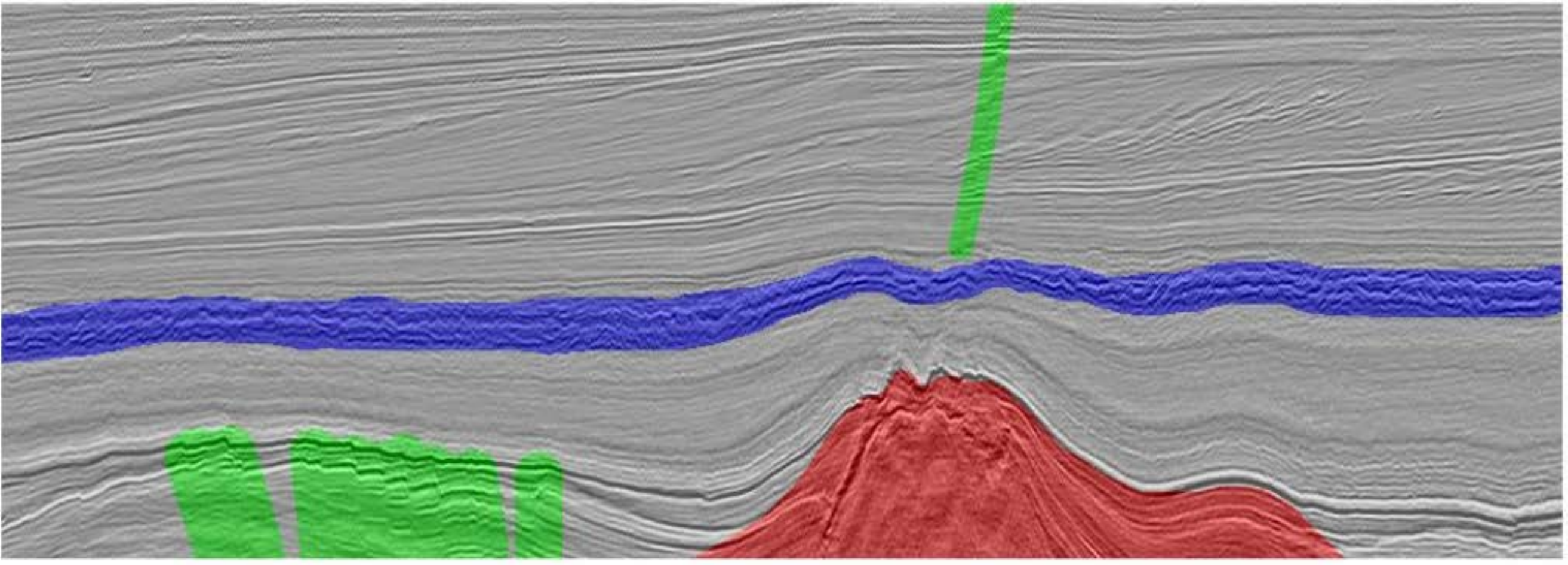} & \\
    (a) & \\
    \includegraphics[width=0.4\linewidth]{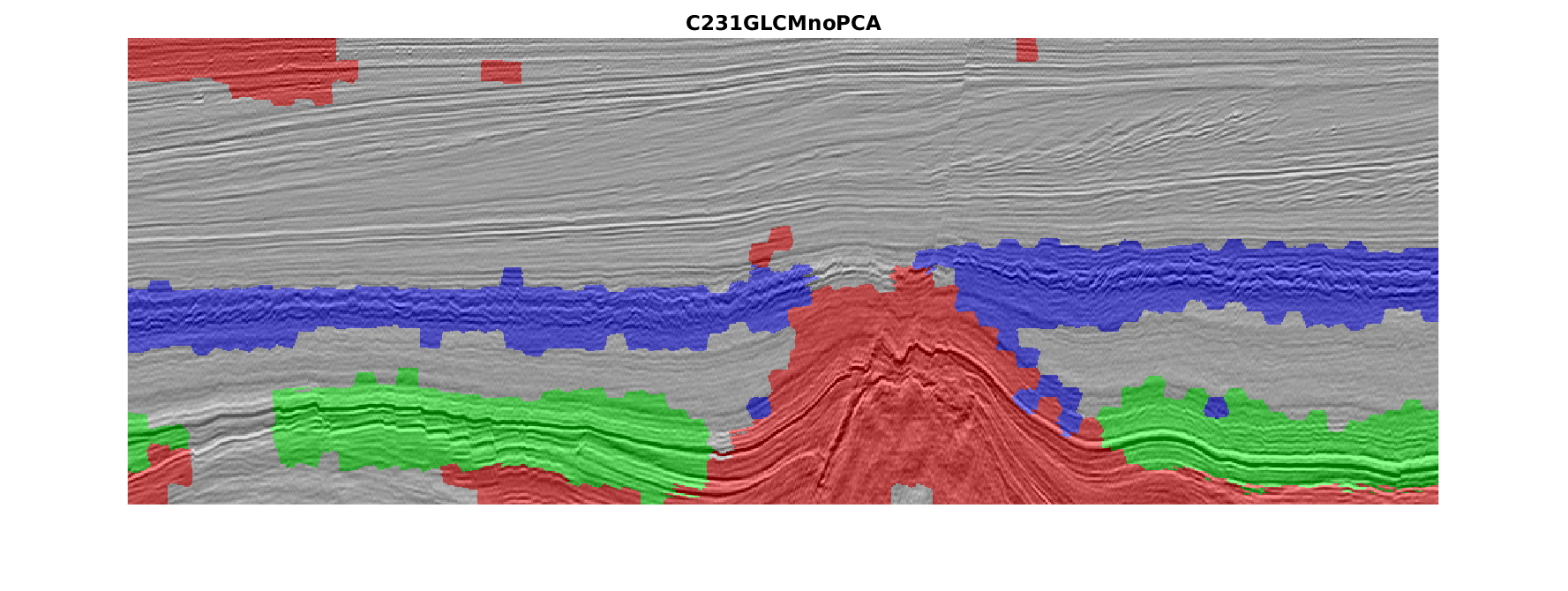} 
    & \includegraphics[width=0.4\linewidth]{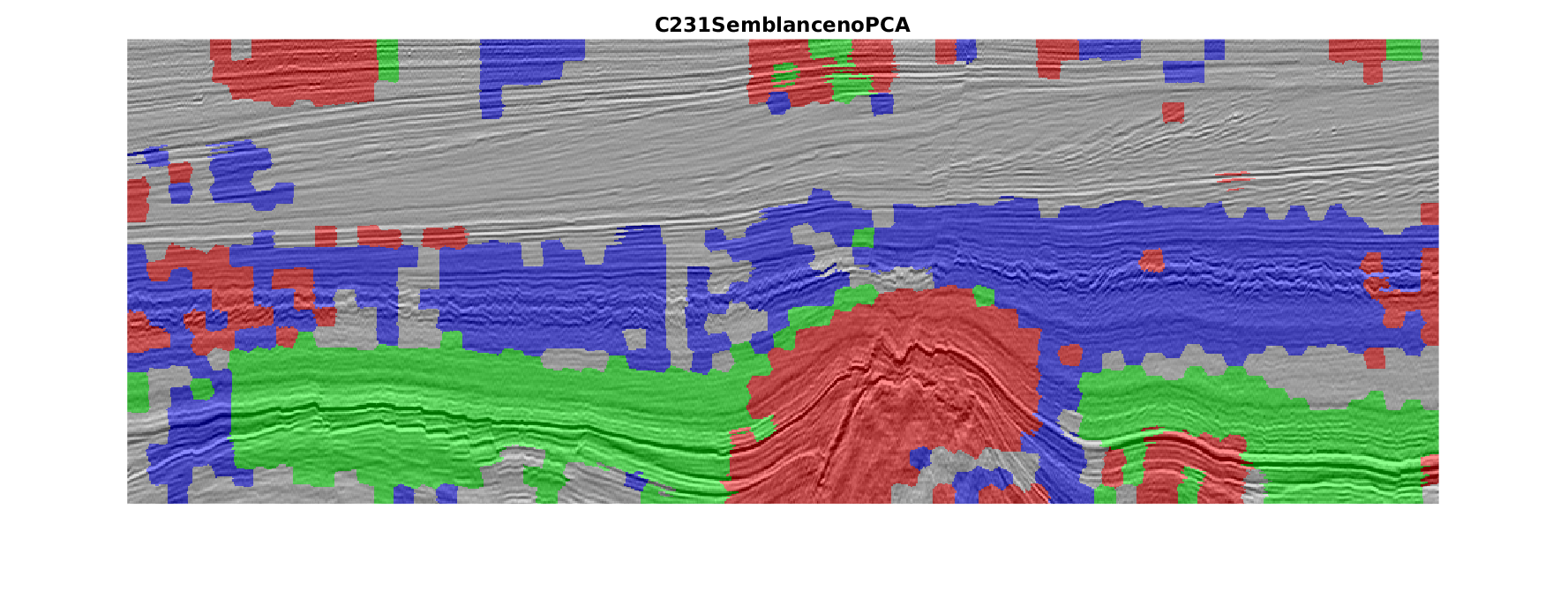} \\
    (b) & (c) \\
	\includegraphics[width=0.4\linewidth]{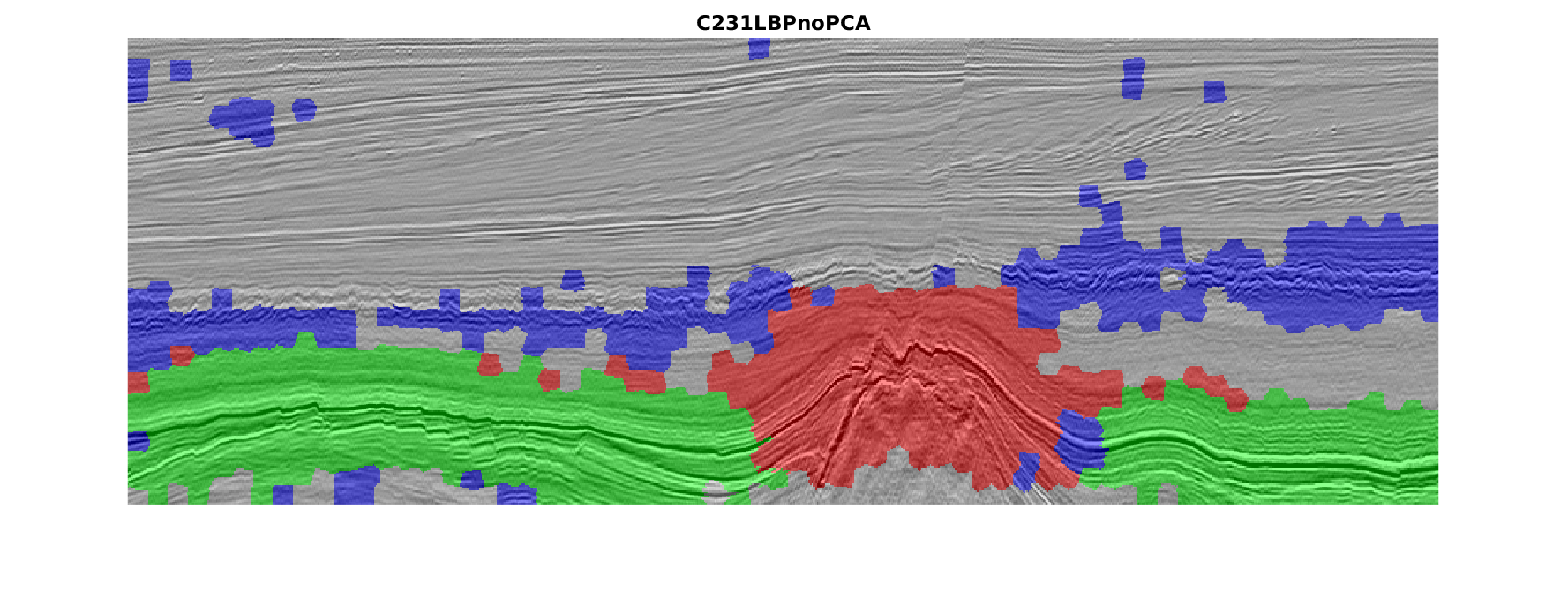} 
	& \includegraphics[width=0.4\linewidth]{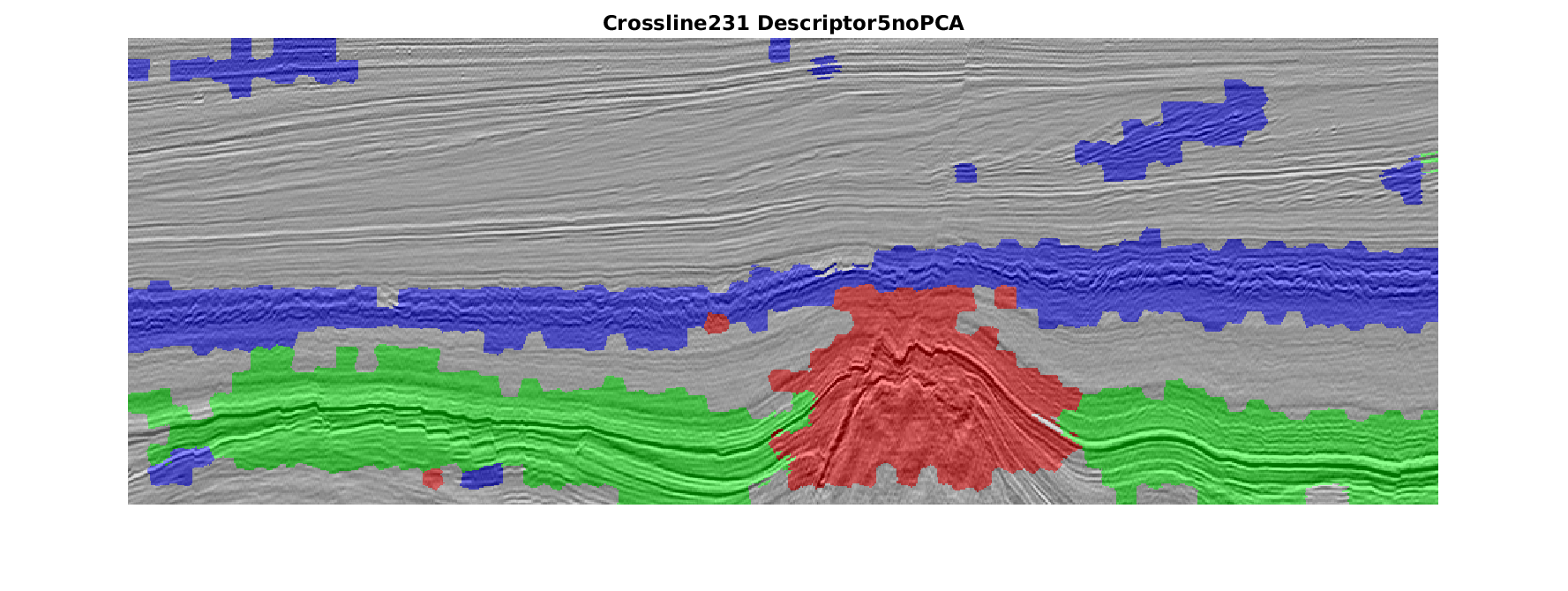} \\
    (d) & (e) \\
	\includegraphics[width=0.4\linewidth]{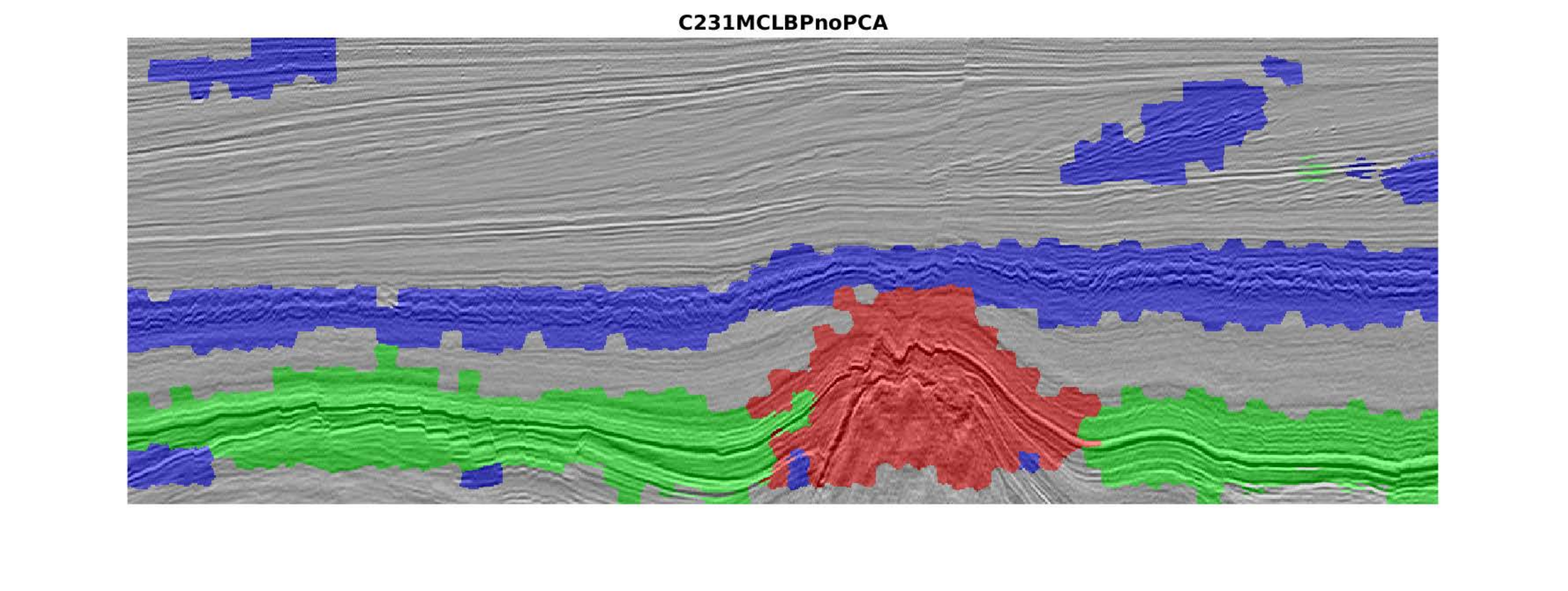} 
	& \includegraphics[width=0.4\linewidth]{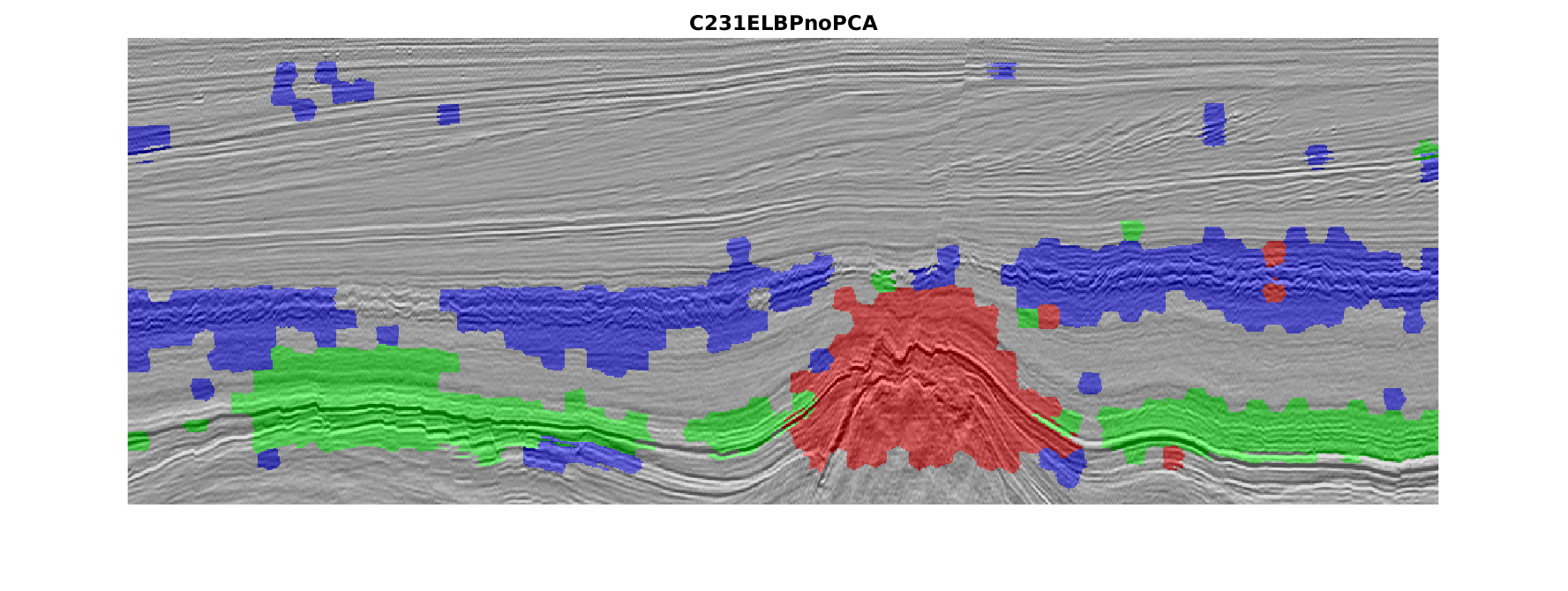} \\
    (f) & (g) \\
	\includegraphics[width=0.4\linewidth]{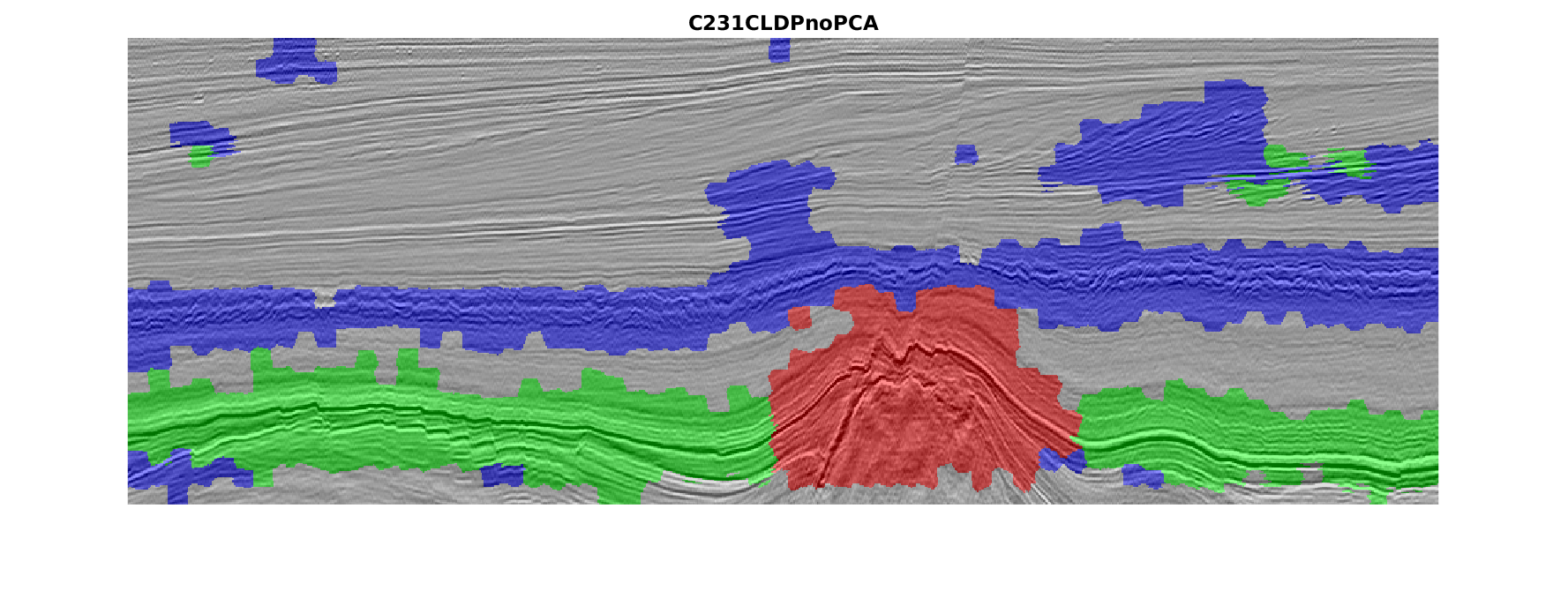} 
	& \includegraphics[width=0.4\linewidth]{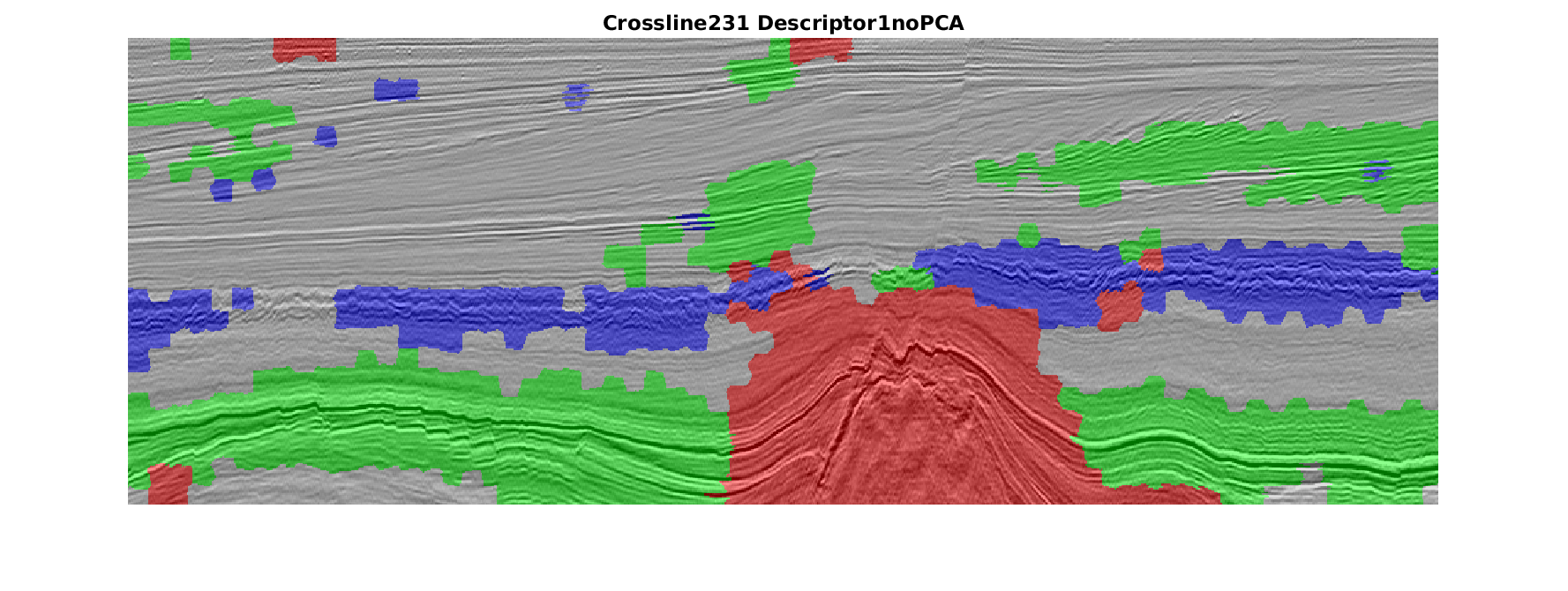} \\
    (h) & (i) \\
\end{tabular}
\caption{Labeling results for the seismic section at crossline 231, in the Netherlands North Sea F3 block database \cite[]{database}, using different texture attributes. The \texttt{Chaotic} class is in blue, \texttt{Faults} is in green, \texttt{Salt Dome} is red, and \texttt{Other} is grey. (a) Manual (manually labeled result overlaid upon the original data); (b) GLCM; (c) Semblance; (d) LBP; (e) CLBP; (f) M-CLBP; (g) ELBP; (h) CLDP; (i) LRI.}
\label{fig:results2}
\end{centering}
\end{figure}

Figure~\ref{fig:results} shows the crossline 281 labeled using each attribute for illustration purpose. The observations are very similar for all four seismic sections being tested. In general, labeling using most of the attributes is able to locate the main structures in presence. The major errors are in the \texttt{Faults} class. This phenomenon is especially true when only one fault exists in the middle of the patch, which is the case for the large fault in the middle of crosslines 211 and 231 (see Figure~\ref{fig:results2}). For GLCM-based labeling, in addition to this single fault, it also misses part of the multiple faults. It is common to label some of the horizons as faults, as well. We believe the reason for the errors in identifying faults lies in the fact that, compared to other structures of interest, \texttt{Faults} are subtle and thin. Patches consisting of faults are typically mixtures of faults and other structures in the neighborhood. Thus, the labeling algorithm can be confused and assign wrong labels. Another factor is that we are also limited by the single \texttt{Faults} exemplar image that was used. To overcome this problem, a solution is to map the patch-level labels into pixel-level labels so that the labeling can be performed at the pixel level. Details of such an approach can be found in a recent work \cite[]{alaudah2017weakly}.

Generally, labeling errors in the results displayed include three types. In some cases, falsely labeled patches are actually connected to correctly labeled structures, appearing as an expanded highlighted area. This kind of labeling errors are usually acceptable if the extended area is within a reasonable range so that an interpreter can still easily determine what structure is being highlighted. Some of the falsely labeled patches are not connected to a correctly labeled structure, but they are grouped together as a few large highlighted areas. Such labeling errors are manageable as they are commonly in small quantities. Thus, an interpreter can probably rule them out quickly. The worst type of labeling errors are those scattered, isolated small patches with wrong labels. Not only are they visually annoying, but they also cause significant inconvenience for an interpreter to examine each one of them.

Specifically, the GLCM-based labeling locates the chaotic structure well. It also identifies the salt dome, but shows some scattered errors at the same time. The semblance-based approach yields a more noisy appearance in the labeled section, especially for the salt dome, which is not desirable. For LBP and its variants, labeling errors for the salt dome are mostly of the first type (i.e., connected errors). Therefore, the labeled salt dome areas generally appear reliable. Regarding the limited scattered errors for the salt dome, the labeling quality is the highest with CLDP, followed by M-CLBP, CLBP, ELBP, and LBP. As to the scattered errors for faults, the LBP-based labeling shows the least of them, while M-CLBP goes to the next and CLDP is associated with the highest. For the chaotic structure, in general, ELBP gives the best labeling quality. Finally, the LRI-based labeling locates the true salt dome better than the LBP-like techniques, capturing more of the salt body. However, at the same time, it shows a little more labeling errors for the salt dome. A major drawback with LRI is that faults in the labeled results exhibit many scattered errors.

\begin{figure}[tbp]
\begin{centering}
\footnotesize
\begin{tabular}{c c}		
	\includegraphics[width=0.45\linewidth]{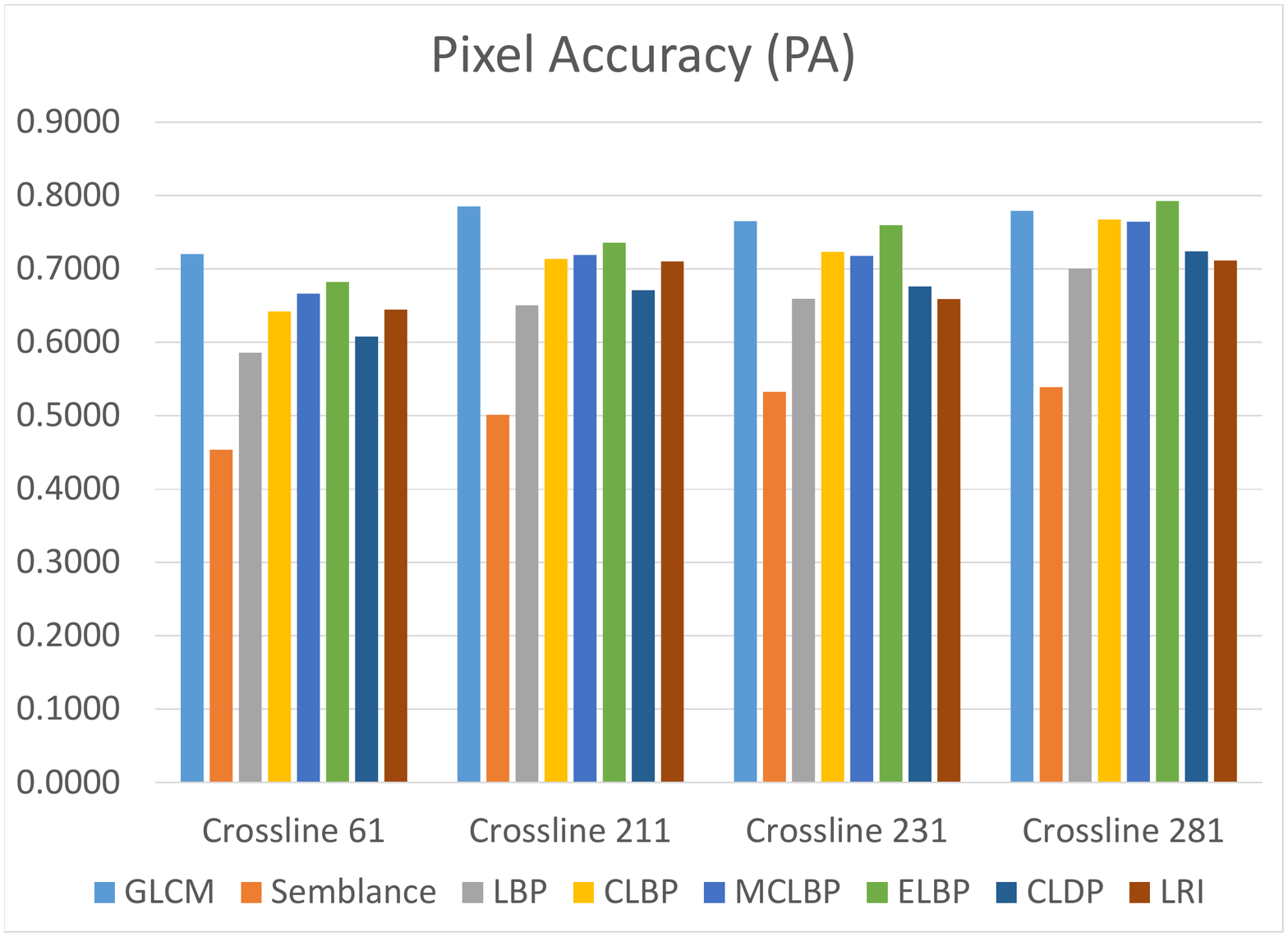}
        & \includegraphics[width=0.45\linewidth]{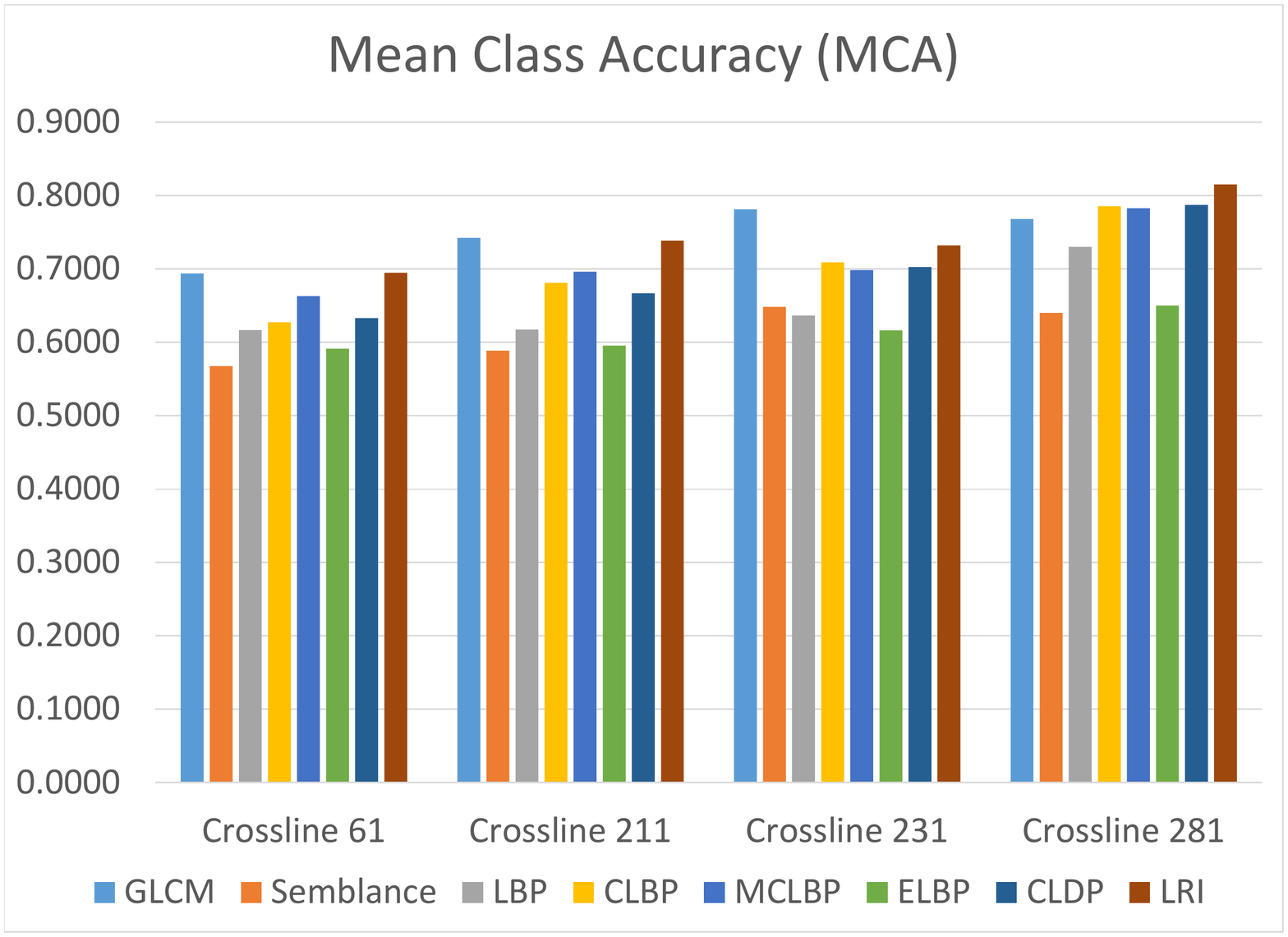} \\
	\includegraphics[width=0.45\linewidth]{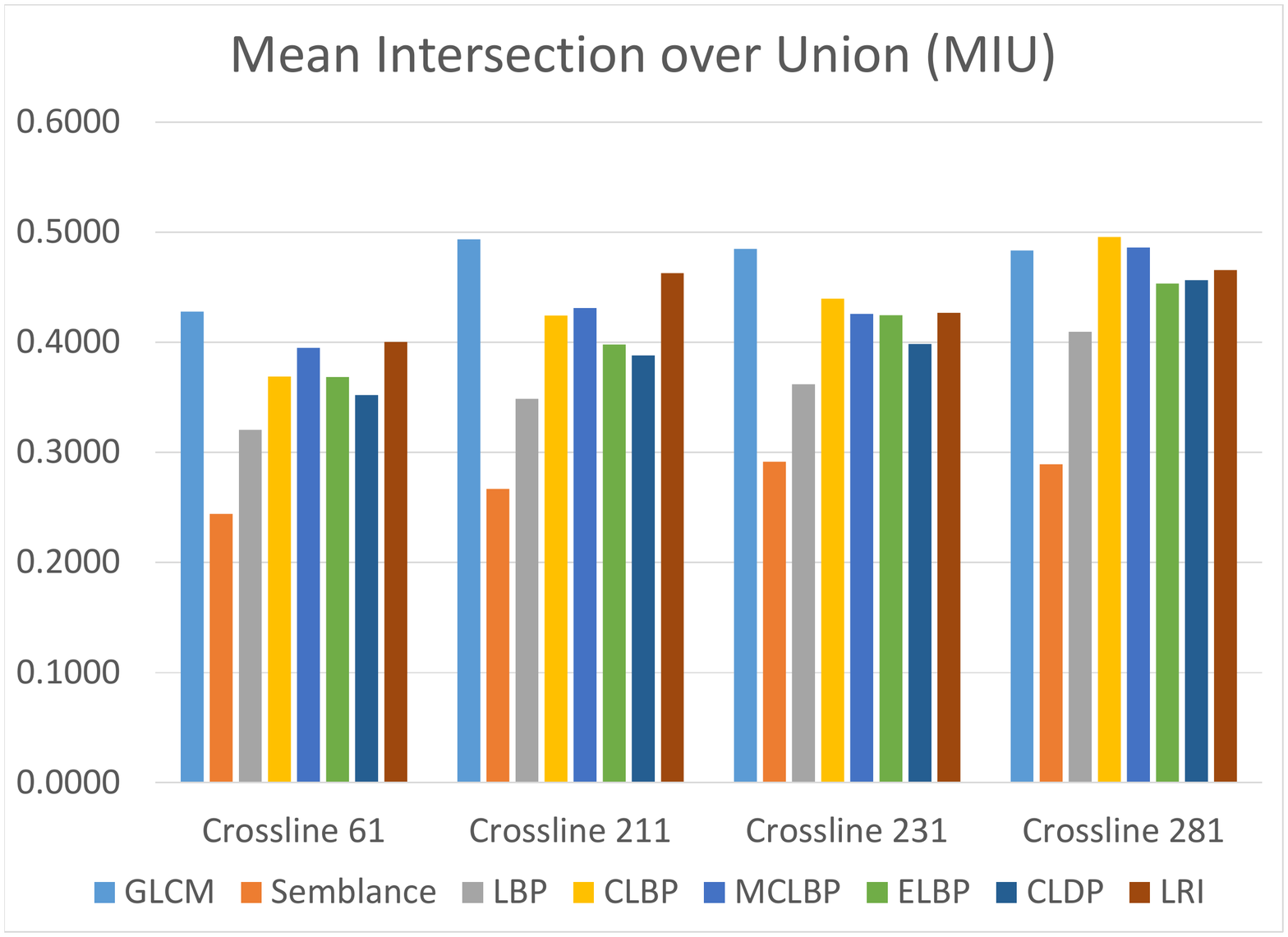}
        & \includegraphics[width=0.45\linewidth]{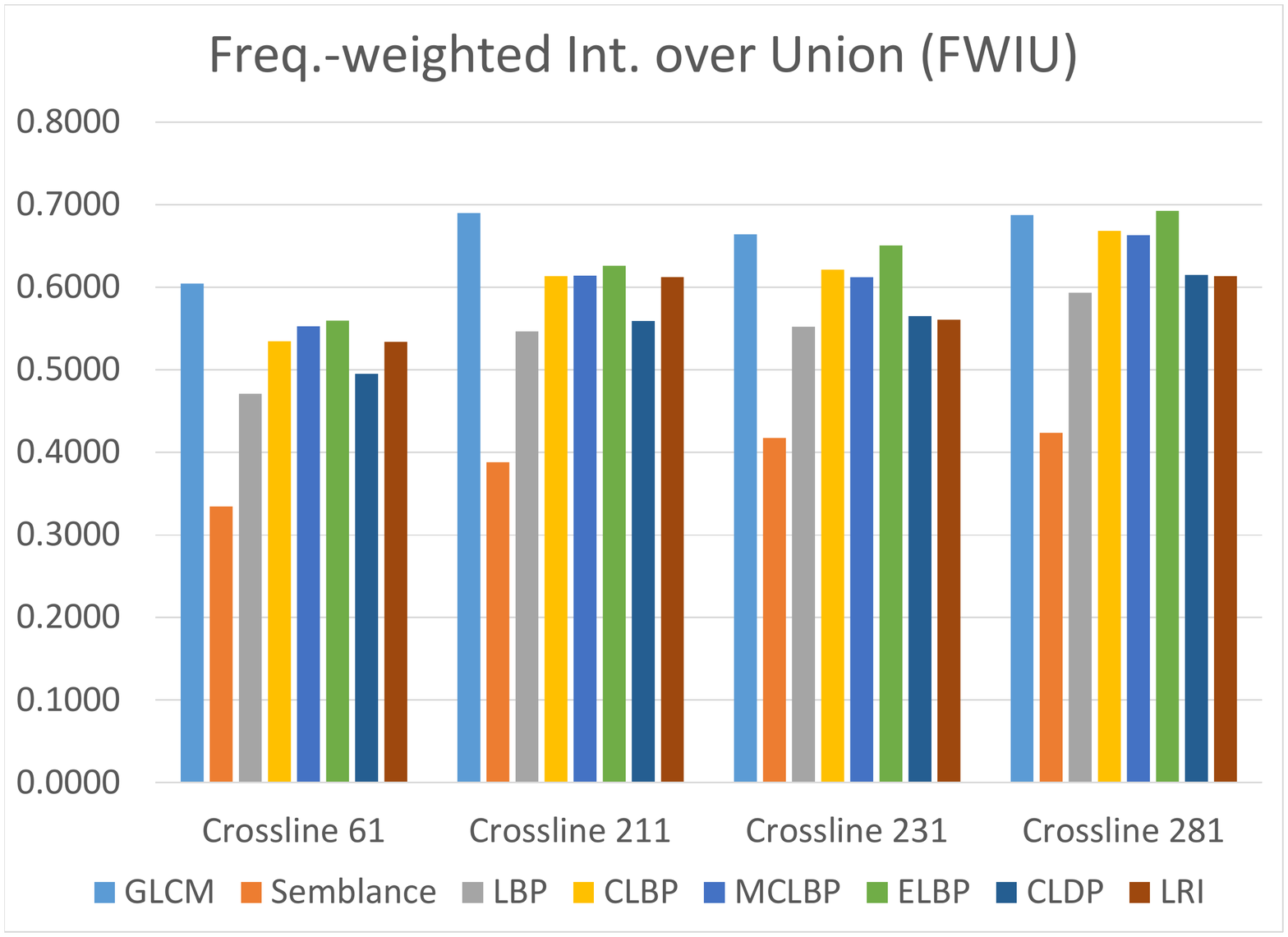} \\
\end{tabular}
\caption{Evaluation of the labeling performance on each seismic section.}
\label{fig:performance}
\end{centering}
\end{figure}

We show objective evaluation results in Table ~\ref{results table} and Figure~\ref{fig:performance}, which match our observation with the actual labeled sections. On average, the two traditional techniques perform at the two extremes, with the GLCM attributes yielding the best results and the semblance being the worst. Among the remaining six local descriptors, ELBP is associated with the highest PA and FWIU, mainly because its background is much less noisy. LRI is the one giving the highest MCA and MIU, indicating that it performs more uniformly across different types of structures. M-CLBP is the only local descriptor that yielded top three performance for all four metrics. As demonstrated in Figure~\ref{fig:performance}, the performance of each attribute is relatively consistent across all four seismic sections being labeled. Considering the very limited manually labeled exemplars that were used, and the challenging nature of the task, the overall performance of the labeling with most texture attributes studied here is very promising.

\begin{table}[tbp]
	\centering
	\caption{Evaluation of the labeling performance, averaged over four different seismic sections, with the top three highlighted for each metric}
	\label{results table}
    \footnotesize
	\begin{tabular}{|c|c|c|c|c|c|c|c|c|}
		\hline
		Attributes & GLCM & SEMB. & LBP & CLBP & M-CLBP & ELBP & CLDP & LRI \\
		\hline
		PA & \textbf{0.7625} & 0.5066 & 0.6491 & 0.7117 & \textbf{0.7171} & \textbf{0.7427} & 0.6698 & 0.6814 \\
		\hline
		MCA & \textbf{0.7465} & 0.6113 & 0.6503 & 0.7008 & \textbf{0.7102} & 0.6134 & 0.6975 & \textbf{0.7453} \\
		\hline
		MIU & \textbf{0.4725} & 0.2729 & 0.3601 & 0.4322 & \textbf{0.4346} & 0.4112 & 0.3989 & \textbf{0.4390} \\
		\hline
		FWIU & \textbf{0.6616} & 0.3910 & 0.5408 & 0.6095 & \textbf{0.6107} & \textbf{0.6322} & 0.5586 & 0.5802 \\
		\hline
	\end{tabular}
\end{table}

\begin{figure}
\begin{center}
\begin{tabular}{c c}
\includegraphics[width=0.5\linewidth]{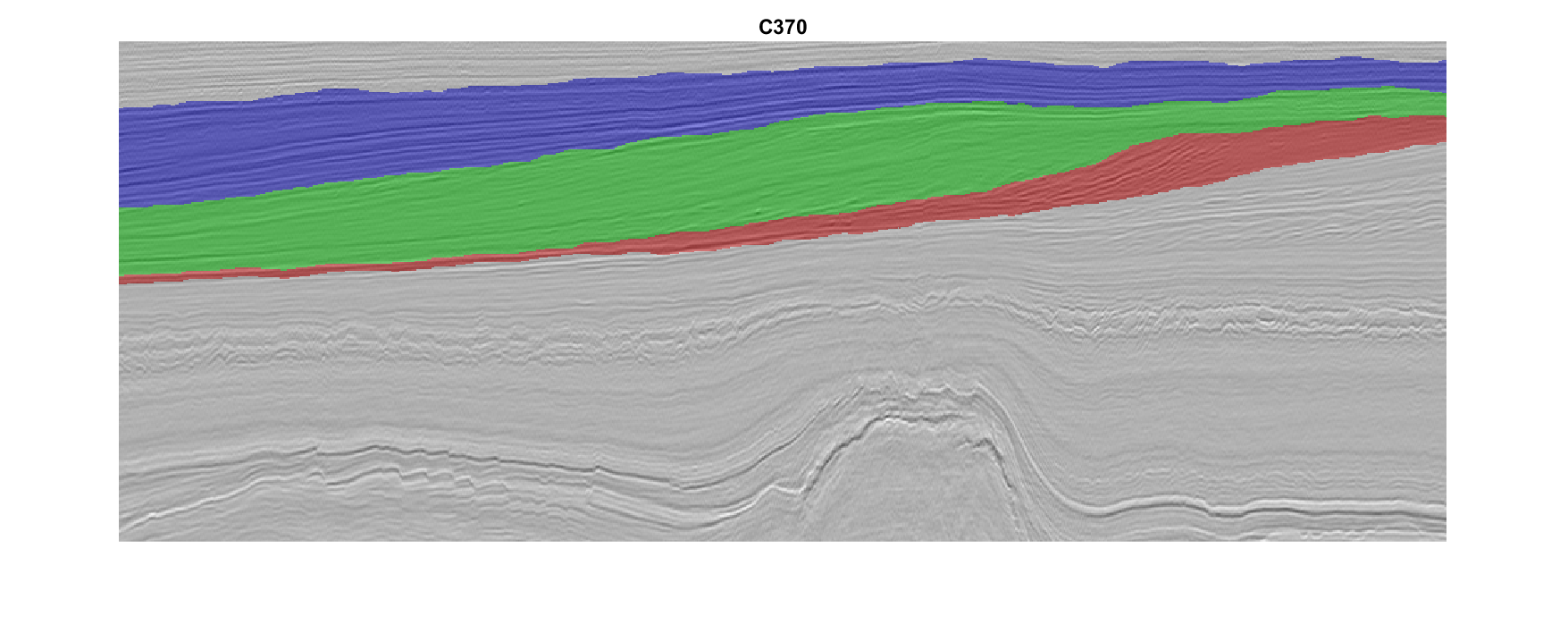} & \includegraphics[width=0.5\linewidth]{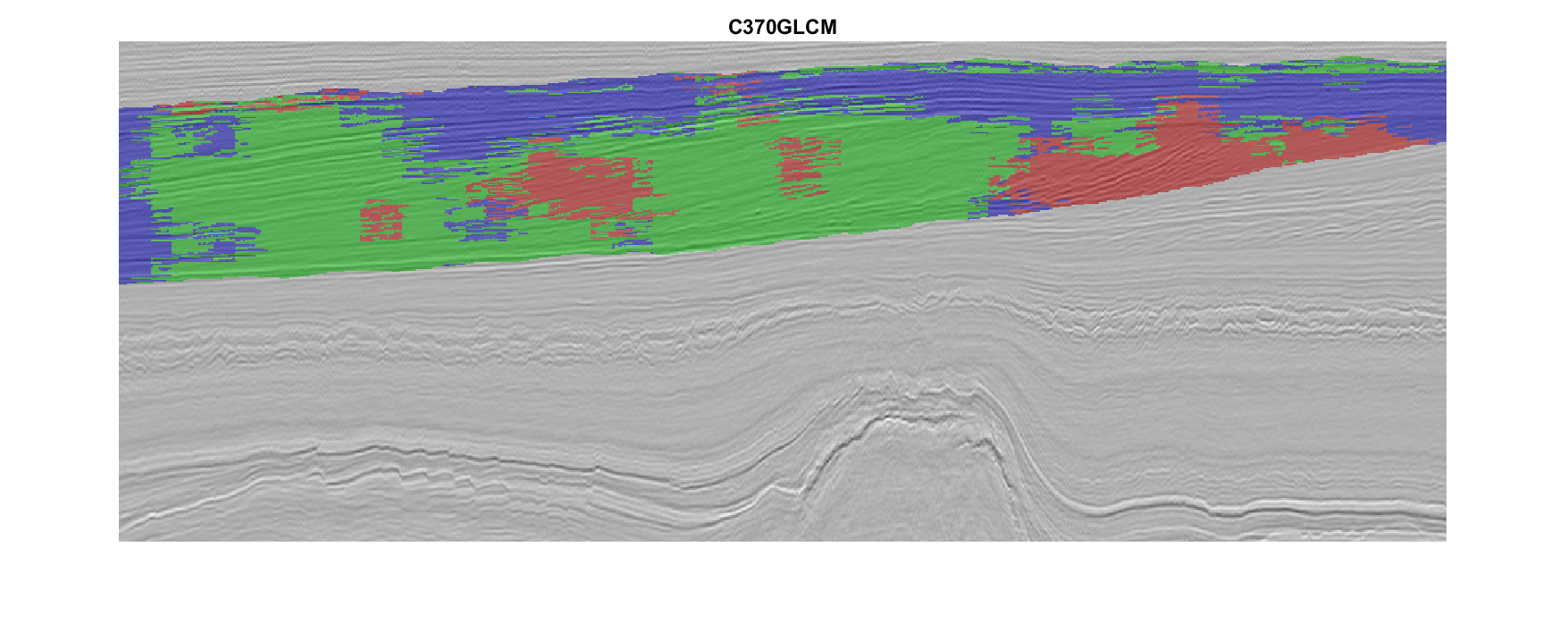} \\
(a) & (b) \\
\includegraphics[width=0.5\linewidth]{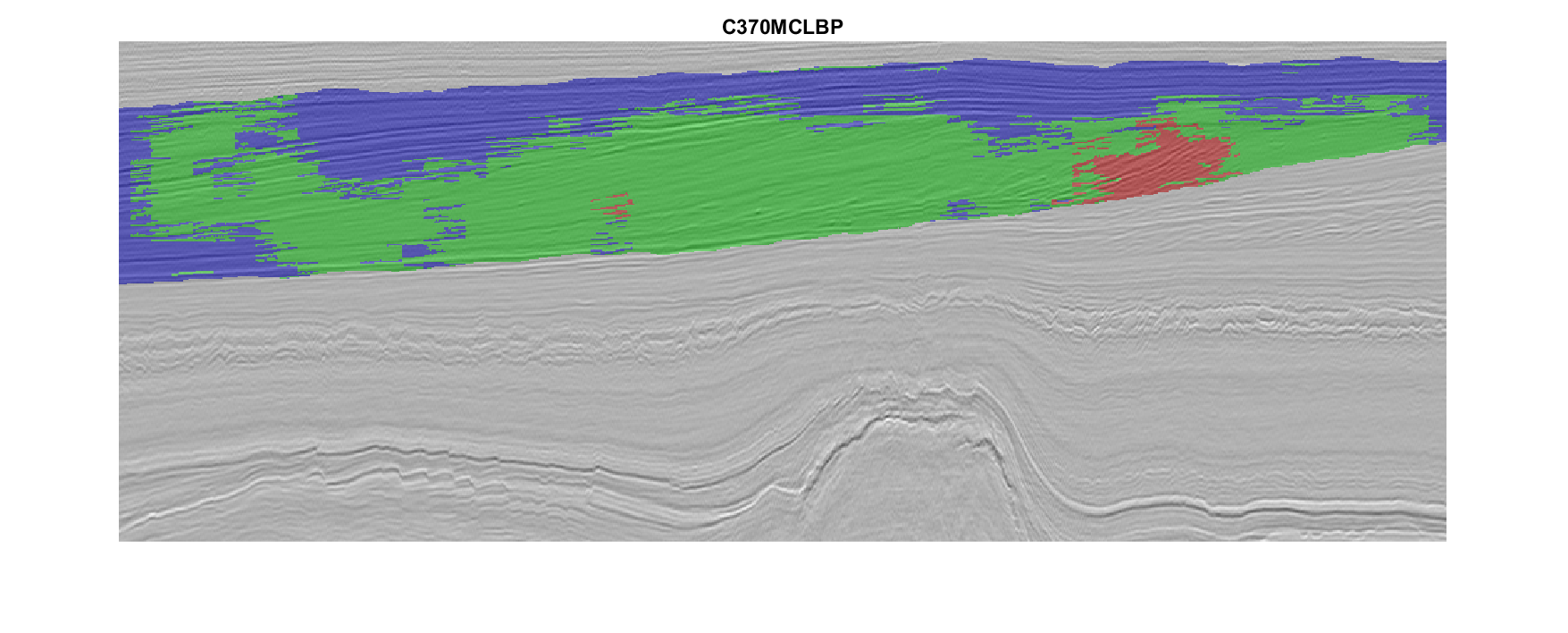} & \includegraphics[width=0.5\linewidth]{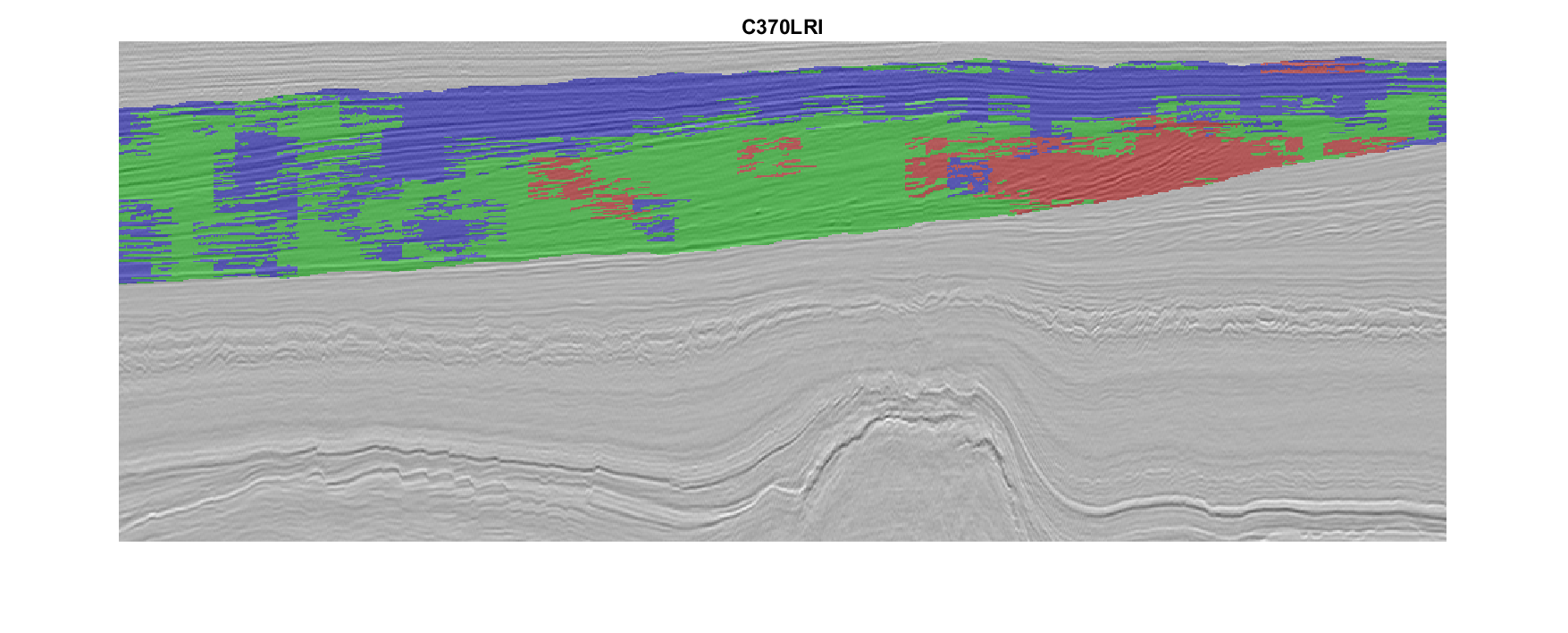}\\
 (c) & (d)\\
\end{tabular}
\end{center}
\caption
{Example results labeling facies using inline 370 of F3 Block. Here, for demonstration purpose, only areas consisting of typical HST, LST, and TST are considered. Training samples were extracted from inline 530. (a) Original inline section with the three facies manually labeled (HST: red, LST: green, and TST: blue); (b) Results using GLCM; (c) Results using M-CLBP; (d) Results using LRI. For the experiment, the patch size adopted was $49\times49$.}
\label{fig:facies}
\end{figure}

In addition to subsurface structure characterization, texture attributes are also important for facies analysis. Naturally, it is of interest to examine if the attributes (i.e., the local descriptors) discussed in this paper can be useful for analyzing facies. Therefore, we applied the framework introduced in this paper to labeling facies instead of structures. We performed a simple labeling experiment to demonstrate the feasibility. In this experiment, we selected three facies in the F3 Block dataset, namely, highstand system tract (HST), lowstand system tract (LST), and transgressive system tract (TST), as described in~\cite[]{illidgeturbidites}. As shown in Figure~\ref{fig:facies}, our labeling tool is capable of producing reasonable results. In particular, the M-CLBP attribute yielded the best labeling output, which is clean and identifies the main locations of each facies very well. Again, this fulfills the purpose of the labeling, which is not to provide accurate delineation but to highlight approximate locations of targets of interest. We also note that the labeling framework is designed towards subsurface structure characterization. Thus, it may need to be adjusted if facies analysis becomes the main application of interest. For example, the segmentation process may be better implemented using the GoT algorithm~\cite[]{shafiq2017texture}. However, further discussion along this direction is beyond the scope of this paper.

%% file: conclusion.tex
In this paper, we conducted a comparative study of a group of spatial texture attributes, including both traditional attributes commonly used in seismic interpretation and local texture descriptors that gained popularity in recent texture image analysis literature. We examined these attributes in a new framework for seismic volume labeling and demonstrated that most of them can be utilized as a generic attribute to characterize different subsurface structures all at the same time. Thus, combined with a suitable interpretation tool such as the labeling tool, they can help provide an initial interpretation with structures of interest being highlighted so that the following interpretation can be expedited.

With respect to the specific attributes discussed in this paper, we found that in the current workflow, none of these attributes can provide satisfactory labeling results for single faults. The subtlety and thinness of such structures are not well captured by the patch-based training. However, we believe the same local descriptors can perform much better for the faults, given that they are able to keep track of local variations at a small scale. The key will be adjusting the workflow to incorporate a pixel-based training so that the classifier is trained with better-suited data such as delineated fault pixels.

The GLCM-based approach yielded the best overall performance in the labeling experiments, which can be attributed to two factors. First, the GLCM attributes are values derived from a histogram, not the histogram itself as with other attributes. Using the derived values can be advantageous in providing a higher-level description of the texture pattern. Second, the GLCM approach combines different types of attributes, creating a more comprehensive representation of various characteristics of the texture pattern. However, creating a GLCM requires a patch of an enough size, which sets such a limit that GLCM attributes are not suitable for small scale structures.

For the purpose of labeling, different attributes can be selected according to the significance of each structure to be located. If salt dome is of the most importance, then CLDP is the best attribute to use for a reliable representation of salt domes. If faults are more important than the others, then LBP and M-CLBP can be considered. If there is no preference for a specific structure, then GLCM, M-CLBP, ELBP and LRI are all good candidates for the labeling system.

There are two purposes of this paper. The first is to introduce and explore the image-based texture attributes, i.e., the local descriptors. The second is to introduce the seismic volume labeling workflow as a useful interpretation tool. However, the labeling tool is not limited to work with the local descriptors only. We believe it is worthwhile to explore its combination with any other kinds of texture attributes. It can also be customized to label specific structures of interest. For example, the labeling tool may be combined with attributes based on textural orientation variations for unconformity detection \cite[]{ringdal2012flow,wu20153d}; it can also be studied in terms of structure tensors \cite[]{bakker2002image,fehmers2003fast,wu2017directional} for better identification of faults and channels \cite[]{wu2017directional2}. We believe these interesting future research will further prove the value of the labeling tool.

%% file: acknowledgements.tex
This work is supported by the Center for Energy and Geo Processing (CeGP) at Georgia Tech and King Fahd University of Petroleum and Minerals (KFUPM). 